\pdfoutput=1

\documentclass[11pt]{article}

\usepackage[preprint]{acl}

\usepackage{times}
\usepackage{latexsym}

\usepackage[T1]{fontenc}
\usepackage[utf8]{inputenc}

\usepackage{microtype}

\usepackage{inconsolata}

\usepackage{graphicx}

\usepackage{url}
\usepackage{booktabs}
\usepackage{amsfonts}
\usepackage{nicefrac}
\usepackage{xcolor}
\usepackage{fancyvrb}
\usepackage{fvextra}

\usepackage{wrapfig}
\usepackage{tikz}
\usepackage{float}
\usepackage{caption}
\usepackage{enumerate}
\usepackage{array}
\usepackage[ruled, linesnumbered, boxed]{algorithm2e}

\usepackage{multirow}
\usepackage{makecell}
\usepackage{bbding}
\usepackage{mathtools}

\usepackage[nameinlink]{cleveref}
\Crefname{figure}{Figure}{Figures}
\crefname{figure}{Figure}{Figures}
\crefname{example}{Example}{Example}
\crefname{theorem}{Theorem}{Theorem}
\crefname{corollary}{Corollary}{Corollary}
\crefname{lemma}{Lemma}{Lemma}
\crefname{proposition}{Proposition}{Proposition}
\crefname{assumption}{Assumption}{Assumption}
\crefname{section}{Section}{Section}
\crefname{algorithm}{Algorithm}{Algorithm}

\usepackage{amsthm,thmtools}
\declaretheorem[name=Theorem,numberwithin=section]{theorem}
\declaretheorem[name=Definition,style=definition]{definition}
\declaretheorem[name=Example,style=definition,numberlike=theorem,qed=\qedsymbol]{example}
\declaretheorem[name=Proposition,numberlike=theorem]{proposition}
\declaretheorem[name=Corollary,numberlike=theorem]{corollary}

\declaretheorem[name=Lemma,numberlike=theorem]{lemma}
\declaretheorem[name=Remark,style=definition,numberwithin=section]{remark}

\graphicspath{{imgs/}{code/results/data_analysis/}{code/results/}}
\usepackage{pgfplots}
\pgfplotsset{compat=1.17}
\usepackage{subcaption}
\usepackage{amssymb}

\newcommand{\R}{\mathbb{R}}
\newcommand{\N}{\mathbb{N}}

\newcommand{\Prob}{\mathbb{P}}

\newcommand{\calS}{\mathcal{S}}
\newcommand{\calA}{\mathcal{A}}

\newcommand{\calM}{\mathcal{M}}
\newcommand{\calH}{\mathcal{H}}

\newcommand{\calT}{\mathcal{T}}

\newcommand{\calV}{\mathcal{V}}
\newcommand{\calB}{\mathcal{B}}

\newcommand{\calI}{\mathcal{I}}
\newcommand{\ind}{\mathbf{1}}  % indicator function

\newcommand{\safe}{\mathsf{Safe}}

\newcommand{\dist}{\mathrm{dist}}

\usepackage{amsmath,amsfonts,bm}

\def\eqref#1{equation~\ref{#1}}
\def\1{\bm{1}}

\DeclareMathAlphabet{\mathsfit}{\encodingdefault}{\sfdefault}{m}{sl}
\SetMathAlphabet{\mathsfit}{bold}{\encodingdefault}{\sfdefault}{bx}{n}

\providecommand{\R}{\mathbb{R}}

\DeclareMathOperator*{\argmin}{arg\,min}

\usepackage{mdframed}

\title{Certified Multi-Turn Robustness for LLM Safety via Compositional Bounds and Safety Persistence}

\author{
\textbf{Yang Liu}$^{1}$ \quad
\textbf{Bin Chong}$^{1,\ast}$ \quad
\textbf{Wenkai Yang}$^{2}$ \quad
\textbf{Shuai Zhang}$^{3}$ \\
\textbf{Yancheng Chen}$^{4}$ \quad
\textbf{Feiyu Han}$^{4}$ \quad
\textbf{GuoZhen}$^{5}$ \quad
\textbf{Cheng Zhang}$^{5}$ \\
\textbf{Huaibing Xie}$^{5}$ \quad
\textbf{Changze Lv}$^{5}$ \quad
\textbf{Shihan Dou}$^{5}$ \quad
\textbf{Pluto Zhou}$^{5}$ \\
$^{1}$Peking University \quad
$^{2}$Renmin University of China \quad
$^{3}$Tsinghua University \\
$^{4}$University of Chinese Academy of Sciences \quad
$^{5}$Tencent Hunyuan \\
$^{\ast}$Corresponding author: \texttt{chongbin@pku.edu.cn}
}

\hypersetup{
  pdftitle={Certified Multi-Turn Robustness for LLM Safety via Compositional Bounds and Safety Persistence},
  pdfauthor={Yang Liu, Bin Chong, Wenkai Yang, Shuai Zhang, Yancheng Chen, Feiyu Han, GuoZhen, Cheng Zhang, Huaibing Xie, Changze Lv, Shihan Dou, Pluto Zhou}
}

\begin{document}
\maketitle
\begin{abstract}
Large language models (LLMs) are vulnerable to multi-turn jailbreak attacks that progressively manipulate conversation context. Existing certified robustness methods are limited to single-turn inputs; naive multi-turn composition yields bounds that degrade exponentially in the number of turns. We introduce \textbf{Multi-Turn Certified Robustness (MTCR)}, a framework that models conversational safety via State-Adversarial MDPs and defines \emph{$k$-turn certified robustness} as the worst-case safety probability across $k$ adversarial turns. MTCR comprises: (i) compositional certification via embedding-space mode decomposition, yielding tighter certified lower bounds than naive multiplication; (ii) $(\alpha,\beta)$-safety persistence, improving the degradation rate from $\underline{p}^{\,k}$ to $\beta^k$ (with $\beta > \underline{p}$) and yielding interpretable horizon estimates; (iii) matching information-theoretic upper bounds establishing tightness; and (iv) a unified algorithm combining these results. Experiments on six LLMs under $\epsilon$-bounded and Crescendo-style attacks confirm that empirical safety consistently exceeds the certified bounds.
\end{abstract}

%==============================================================================
% SECTION 1: INTRODUCTION
%==============================================================================
\section{Introduction}
\label{sec:introduction}

The deployment of large language models (LLMs) in conversational applications has raised safety concerns, particularly regarding \emph{jailbreak attacks}: carefully crafted inputs designed to bypass safety mechanisms and elicit harmful content \citep{zou2023universal,wei2024jailbroken}. Although substantial progress has been made in defending against single-turn attacks, \textbf{multi-turn jailbreak attacks} that progressively manipulate conversation context across multiple interactions pose a distinct challenge \citep{russinovich2024crescendo,li2024mhj}.

Recent empirical studies reveal the severity of this threat. The Crescendo attack \citep{russinovich2024crescendo} achieves near-perfect success rates against production LLMs using fewer than 10 turns; X-Teaming \citep{xteaming2025} attains 96--98\% success rates against state-of-the-art models; human red-teamers consistently exceed 70\% success against deployed defenses \citep{li2024mhj}. These findings indicate that \emph{defenses robust against single-turn attacks offer little protection in multi-turn settings}.

On the theoretical side, existing certified robustness methods (randomized smoothing \citep{robey2024smoothllm}, erase-and-check \citep{kumar2024certifying}, knapsack-based certification \citep{chen2025worstcase}) are limited to single-input settings. A naive extension yields certified lower bounds that degrade {exponentially} in the number of turns: if each turn has worst-case certified safety probability $p$, the $k$-turn bound under multiplicative composition is merely $p^k$, which becomes negligible for moderate $k$. This degradation arises because adaptive adversaries observe responses and craft subsequent inputs (a sequential game), accumulated context compounds adversarial influence, and autoregressive generation amplifies early-token perturbations.

\paragraph{Contributions.} We develop \textbf{Multi-Turn Certified Robustness (MTCR)}, a certification framework grounded in SA-MDP theory, with three contributions:

\textbf{(1)} We model multi-turn conversations as State-Adversarial MDPs and define {$k$-turn certified robustness}. We develop {compositional certification} via embedding-space mode decomposition: certifying intra-mode safety and inter-mode transitions separately yields bounds tighter than naive multiplication (Corollary~\ref{cor:improvement}). A unified algorithm (Algorithm~\ref{alg:mtcr}) combines compositional and persistence bounds to compute the certified guarantee in practice.

\textbf{(2)} We formalize {$(\alpha,\beta)$-safety persistence}, a structural property under which the robustness bound improves from $\underline{p}^{\,k}$ to $\beta^k$ (when $\beta > \underline{p}$), with a linear characterization $1 - k(1-\beta)$ that yields interpretable horizon estimates (Theorem~\ref{thm:persistence}). We prove information-theoretic upper bounds showing our compositional approach is tight for non-overlapping decompositions, and impossibility results for systems that lack structural assumptions.

\textbf{(3)} We evaluate MTCR on six production LLMs. The certified bounds provide formal guarantees against $\epsilon$-bounded adversaries; empirical safety consistently exceeds the bound under both static ($\epsilon$-ball) and Crescendo-style adaptive attacks, confirming that the bounds are not violated in practice.

\begin{figure*}[t]
\centering
\includegraphics[width=\linewidth]{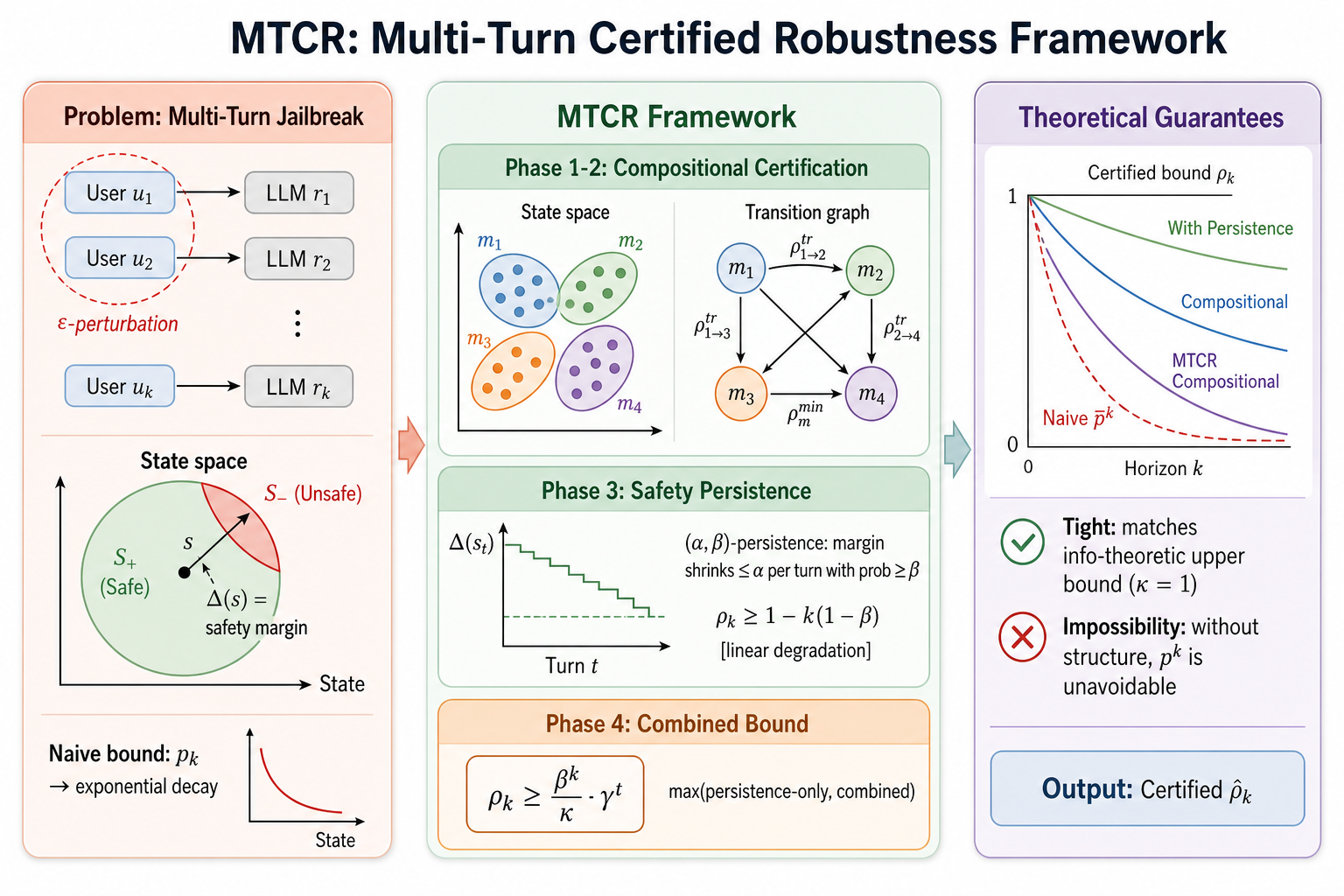}
\caption{Overview of MTCR. \textbf{Left}: Multi-turn conversations modeled as SA-MDPs; naive bound $\underline{p}^{\,k}$ degrades exponentially. \textbf{Center}: MTCR combines compositional certification (intra-mode + transition safety) with $(\alpha,\beta)$-persistence into a unified bound. \textbf{Right}: Matching upper bounds establish tightness ($\kappa{=}1$). Experiments on six LLMs confirm empirical safety exceeds the certified bounds.}
\label{fig:framework}
\end{figure*}
%==============================================================================
% SECTION 2: PROBLEM FORMULATION
%==============================================================================
\section{Problem Formulation: Multi-Turn Safety as SA-MDP}
\label{sec:framework}

We model multi-turn conversations as State-Adversarial MDPs (SA-MDPs). The dialogue history $h_t = (u_1, r_1, \ldots, u_t, r_t)$ maps to a conversational state $s_t = \phi(h_t) \in \calS \subseteq \R^d$ via an embedding $\phi$ (e.g., LLM hidden representations). An adversary selects user messages $u_t$ from an $\epsilon$-perturbation ball $\calB_\epsilon$ around a reference input; the LLM responds according to policy $\pi(\cdot|s,u)$. States evolve as $s_t = T(s_{t-1}, u_t, r_t)$. A safety predicate $\safe: \calS \to \{0,1\}$ partitions the state space into a safe region $\calS_+$ and an unsafe region $\calS_-$, with safety margin $\Delta(s) = \dist(s, \calS_-)$. We assume $\calS$ is bounded, $\calS_+$ is open (so $\Delta(s) > 0$ for all $s \in \calS_+$), and $\calS_+$ has non-empty interior with $\delta_{\min}$ denoting the maximum inscribed ball radius. Full formal definitions appear in Appendix~\ref{app:formal_defs}.

\begin{definition}[$k$-Turn Certified Robustness]
\label{def:k_turn}
Given initial state $s_0 \in \calS_+$, LLM policy $\pi$, perturbation budget $\epsilon$, and horizon $k \in \N$, the {$k$-turn certified robustness} $\rho_k(s_0, \pi, \epsilon)$ is:
\begin{equation}
\inf_{\nu \in \Pi_\epsilon} \Prob_{\substack{u_t \sim \nu(\cdot|s_{t-1}) \\ r_t \sim \pi(\cdot|s_{t-1},u_t)}}\left[\bigwedge_{t=1}^{k} \safe(s_t) \,\bigg|\, s_0\right],
\end{equation}
where states evolve according to $s_t = T(s_{t-1}, u_t, r_t)$.
\end{definition}

The quantity $\rho_k(s_0, \pi, \epsilon)$ captures the worst-case probability of maintaining safety across $k$ turns against any adaptive adversary (one that observes $s_{t-1}$ before choosing $u_t$). A natural certification strategy is to compose single-turn safety bounds multiplicatively. For each state $s$, let $p(s,\epsilon) = \inf_{u \in \calB_\epsilon} \Prob[\safe(T(s,u,r))]$ denote the certified probability of remaining safe after one adversarial turn. This composition yields the following bound.

\begin{proposition}[Naive Multiplicative Bound]
\label{prop:naive}
For any $s_0 \in \calS_+$:
\begin{equation}
    \rho_k(s_0, \pi, \epsilon) \geq \left(\inf_{s \in \calS_+} p(s, \epsilon)\right)^k = \underline{p}^{\,k},
\end{equation}
where $\underline{p} = \inf_{s \in \calS_+} p(s, \epsilon)$ is the worst-case per-turn certified safety. If $\underline{p} < 1$, this bound vanishes exponentially in $k$.
\end{proposition}

\begin{example}[Vacuity of Naive Bound]
\label{ex:vacuous}
With per-turn safety $\underline{p} = 0.95$: $\rho_{10} \geq 0.60$, $\rho_{20} \geq 0.36$, $\rho_{50} \geq 0.08$. For safety-critical applications requiring $\rho_k \geq 0.9$, this permits at most $k \leq 2$ turns.
\end{example}

\paragraph{Scope of certification.} The certified guarantee applies to adversaries constrained within an $\epsilon$-ball (e.g., character-level edit distance) around reference inputs. Real attacks such as Crescendo operate at the semantic level; while our experiments report empirical safety against such attacks, the \emph{formal certification} covers only $\epsilon$-bounded perturbations. The SA-MDP framework is agnostic to the perturbation metric and extends to richer threat models once compatible per-turn certification oracles become available.

Without additional structure, the multiplicative composition provides little practical guarantee for conversations beyond a few turns. This motivates identifying structural conditions under which tighter bounds can be achieved.

%==============================================================================
% SECTION 3: MULTI-TURN CERTIFIED ROBUSTNESS (Method)
%==============================================================================
\section{Multi-Turn Certified Robustness}
\label{sec:method}

We extend single-turn certification to multi-turn conversations by exploiting mode structure in embedding space and safety margin evolution. The framework has four parts: compositional bounds via mode decomposition (\S\ref{sec:compositional}), safety persistence characterizing margin evolution (\S\ref{sec:persistence}), information-theoretic upper bounds establishing tightness (\S\ref{sec:lower_bounds}), and a unified algorithm (\S\ref{sec:algorithm}).

\subsection{Compositional Certification}
\label{sec:compositional}

Conversations exhibit \emph{mode structure} in embedding space that can be exploited for tighter certification. A {mode decomposition} $\calM = \{m_1, \ldots, m_M\}$ covers $\calS_+$ with bounded overlap $\kappa = \max_s |\{i : s \in \calS_{m_i}\}|$. For each mode $m$ we define {intra-mode certified safety} $\rho_m^{\mathrm{in}}(\epsilon)$ (worst-case probability of staying safe and in $m$) and for edges $(m_i,m_j)$ in the transition graph we define {transition safety} $\rho_{i \to j}^{\mathrm{tr}}(\epsilon)$. A mode trajectory $\sigma$ over $k$ turns has $n_j(\sigma)$ intra-mode turns in $m_j$ and $\tau(\sigma)$ transitions. Full definitions appear in Appendix~\ref{app:formal_defs}.

\begin{theorem}[Compositional Certification Bound]
\label{thm:compositional}
Let $\calM = \{m_1, \ldots, m_M\}$ be a mode decomposition with overlap $\kappa = \kappa(\calM)$. For any $k$-turn conversation with mode trajectory $\sigma \in \Sigma_k(\calM)$:
\begin{align*}
\rho_k(s_0, \pi, \epsilon) \geq \frac{1}{\kappa} &\cdot \prod_{j=1}^{M} \left(\rho_{m_j}^{\mathrm{in}}(\epsilon)\right)^{n_j(\sigma)} \\
&\cdot \prod_{(i,j) \in \mathrm{Trans}(\sigma)} \rho_{i \to j}^{\mathrm{tr}}(\epsilon)
\end{align*}
where $\mathrm{Trans}(\sigma) = \{(\sigma(t), \sigma(t+1)) : \sigma(t) \neq \sigma(t+1)\}$ is the multiset of transitions. Taking the infimum over feasible trajectories:
\begin{align*}
\rho_k(s_0, \pi, \epsilon) \geq \frac{1}{\kappa} \cdot &\inf_{\sigma \in \Sigma_k(\calM, s_0)} \bigg[\prod_{j=1}^{M} \left(\rho_{m_j}^{\mathrm{in}}\right)^{n_j(\sigma)} \\
&\cdot \prod_{(i,j) \in \mathrm{Trans}(\sigma)} \rho_{i \to j}^{\mathrm{tr}}\bigg]
\end{align*}
where $\Sigma_k(\calM, s_0) \subseteq \Sigma_k(\calM)$ are trajectories starting from a mode containing $s_0$.
\end{theorem}

The following corollary quantifies when the compositional bound strictly improves upon the naive multiplicative bound.

\begin{corollary}[Improvement over Naive Composition]
\label{cor:improvement}
Let $\underline{p} = \inf_s p(s,\epsilon)$ be the naive per-turn bound. Suppose the mode decomposition satisfies:
\begin{enumerate}
    \item Intra-mode safety: $\rho_{m_j}^{\mathrm{in}} \geq 1 - \delta$ for small $\delta > 0$
    \item Inter-mode transition safety: $\rho_{i \to j}^{\mathrm{tr}} \geq \gamma$ for some $\gamma \in (0,1)$
    \item Sparse transitions: trajectory $\sigma$ has at most $\tau$ transitions
\end{enumerate}
Then the compositional bound is:
\begin{equation}
    \rho_k^{\mathrm{comp}} \geq \frac{1}{\kappa}(1-\delta)^{k-\tau} \cdot \gamma^\tau
\end{equation}
This exceeds the naive bound $\underline{p}^{\,k}$ when:
\begin{equation}
    \tau < \frac{k \log(1-\delta) - k\log\underline{p} - \log\kappa}{\log(1-\delta) - \log\gamma}
\end{equation}
\end{corollary}

\begin{example}[Numerical Comparison]
\label{ex:numerical}
Under parameters $k{=}20$, $\tau{=}3$, $\underline{p}{=}0.9$, $\delta{=}0.02$, $\gamma{=}0.7$, $\kappa{=}1$:
\begin{align*}
    \rho_{20}^{\mathrm{naive}} &\geq 0.9^{20} \approx 0.122 \\
    \rho_{20}^{\mathrm{comp}} &\geq 0.98^{17} \cdot 0.7^3 \approx 0.708 \cdot 0.343 \approx 0.243
\end{align*}
The compositional certified lower bound is approximately \textbf{twice} the naive lower bound. Note: improvement depends on $k$ and $\tau$; when $k$ is small, transition overhead may dominate (see Appendix~\ref{app:comp_gain}).
\end{example}

Computing the compositional bound requires finding the worst-case trajectory over the mode transition graph. This can be done efficiently, as the following proposition shows.

\begin{proposition}[Worst-Case Trajectory Computation]
\label{prop:dp}
The trajectory infimum in Theorem~\ref{thm:compositional} can be computed by dynamic programming in $O(M^2 k)$ time.
\end{proposition}

\subsection{Safety Persistence}
\label{sec:persistence}

The compositional bound still depends on the number of transitions $\tau$. We next define \emph{safety persistence}, a property that enables stronger guarantees by characterizing how safety margins evolve.

\begin{definition}[$(\alpha,\beta)$-Safety Persistence]
\label{def:persistence}
A conversational system $(\calS, T, \pi, \safe)$ exhibits \textbf{$(\alpha,\beta)$-safety persistence} for $\alpha \in [0,1)$ and $\beta \in (0,1]$ if for all safe states $s \in \calS_+$ with safety margin $\Delta(s) > 0$:
\begin{equation}
    \Prob_{u \sim \nu,\, r \sim \pi}\!\left[
    \begin{gathered}
    \safe(s') = 1 \\
    {}\land\ \Delta(s') \geq (1-\alpha)\Delta(s)
    \end{gathered}
    \right] \geq \beta
\end{equation}
where $s' = T(s, u, r)$ and $\nu \in \Pi_\epsilon$ is any adversarial policy.
\end{definition}
With probability at least $\beta$, the safety margin retains at least a $(1{-}\alpha)$ fraction of its previous value per turn; well-aligned LLMs typically satisfy this property. Under safety persistence, we obtain the following guarantee.

\begin{theorem}[Degradation under Safety Persistence]
\label{thm:persistence}
Suppose the system exhibits $(\alpha, \beta)$-safety persistence. Let $s_0 \in \calS_+$ have initial margin $\Delta_0 = \Delta(s_0) > 0$. Then:
\begin{equation}
    \rho_k(s_0, \pi, \epsilon) \geq \beta^k
\end{equation}
When $\beta > \underline{p}$ (i.e., the persistence guarantee exceeds the worst-case per-turn bound), we have $\beta^k > \underline{p}^{\,k}$, strictly improving upon the naive bound.

Additionally, a union bound yields an interpretable (though numerically weaker) characterization:
\begin{equation}
    \rho_k(s_0, \pi, \epsilon) \geq 1 - k(1-\beta)
\end{equation}
This formula provides a sufficient condition for safety: the system remains safe with probability $\geq 1-\xi$ whenever $k \leq \xi/(1-\beta)$, enabling direct horizon estimation.
\end{theorem}

\begin{corollary}[Effective Horizon Estimation]
\label{cor:horizon}
Under $(\alpha,\beta)$-safety persistence, to maintain $\rho_k \geq 1 - \xi$ for target $\xi \in (0,1)$, the linear characterization gives the sufficient horizon bound:
\begin{equation}
    k_{\max} = \left\lfloor\frac{\xi}{1-\beta}\right\rfloor
\end{equation}
compared to $k_{\max} = \lfloor\log(1-\xi)/\log\underline{p}\rfloor$ under naive composition. For example, with $\beta{=}0.98$ and $\xi{=}0.1$: the persistence-based estimate gives $k_{\max} = 5$, whereas $\underline{p}{=}0.9$ gives $k_{\max} = 1$.
\end{corollary}

Safety persistence can be verified from Lipschitz continuity of the transition dynamics and concentration properties of the LLM output; the following proposition gives sufficient conditions.

\begin{proposition}[Persistence from Lipschitz Continuity]
\label{prop:lipschitz}
Suppose:
\begin{enumerate}
    \item $T$ is $L_s$-Lipschitz in state: $\|T(s,u,r) - T(s',u,r)\| \leq L_s\|s-s'\|$
    \item $T$ is $L_u$-Lipschitz in input: $\|T(s,u,r) - T(s,u',r)\| \leq L_u \cdot d(u,u')$
    \item $T$ is $L_r$-Lipschitz in response: $\|T(s,u,r) - T(s,u,r')\| \leq L_r \cdot d_r(r,r')$ for some metric $d_r$
    \item The safe region satisfies: $\calS_+$ is convex with $\dist(s, \calS_-) \geq \delta$ for $s$ in the $\delta$-interior
    \item LLM responses satisfy: $\Prob[\|r - r^*\| \leq \eta] \geq p_0$ for some nominal response $r^*$
    \item Nominal drift is bounded: $\|T(s,u^*,r^*) - s\| \leq \delta_T$ for all $s \in \calS_+$
\end{enumerate}
Then for all states $s$ in the $\delta$-interior of $\calS_+$ (i.e., $\Delta(s) \geq \delta$), the $(\alpha, \beta)$-persistence condition (Definition~\ref{def:persistence}) holds with:
\begin{equation}
\begin{aligned}
    \alpha &= \frac{L_u\epsilon + L_r\eta + \delta_T}{\delta}, \\
    \beta &= p_0 \cdot \ind[L_u\epsilon + L_r\eta + \delta_T < \delta].
\end{aligned}
\end{equation}
The indicator function reflects a strict sufficient condition; when $L_u\epsilon + L_r\eta + \delta_T \geq \delta$, it yields $\beta = 0$, making the bound vacuous. This occurs in many practical settings where the perturbation budget and nominal drift are large relative to the safety margin. Proposition~\ref{prop:lipschitz} serves as a theoretical grounding showing \emph{when} persistence holds from first principles; in our experiments, persistence parameters are instead estimated empirically via sampling (Algorithm~\ref{alg:mtcr} Phase~1; see Appendix~\ref{app:persistence_estimation}).
\end{proposition}

\subsection{Tightness and Impossibility Results}
\label{sec:lower_bounds}

We establish that our compositional bounds are tight. The following theorem gives an information-theoretic upper bound via a matching construction, showing that the compositional product cannot be improved for non-overlapping decompositions.

\begin{theorem}[Information-Theoretic Upper Bound on Certification]
\label{thm:upper_bound}
For any mode decomposition $\calM$ and any trajectory $\sigma$, there exist transition dynamics $T$ and adversarial strategies $\nu^*$ such that:
\begin{align*}
\rho_k(s_0, \pi, \epsilon) \leq \prod_{j=1}^M &\left(\rho_{m_j}^{\mathrm{in}}\right)^{n_j(\sigma)} \\
&\cdot \prod_{(i,j) \in \mathrm{Trans}(\sigma)} \rho_{i \to j}^{\mathrm{tr}}
\end{align*}
The compositional product is also an upper bound for a worst-case system matching the mode-level safety parameters.
\end{theorem}

\begin{corollary}[Near-Optimality of Compositional Bounds]
\label{cor:optimal}
For non-overlapping mode decompositions ($\kappa = 1$), our compositional lower bound (Theorem~\ref{thm:compositional}) matches the information-theoretic upper bound (Theorem~\ref{thm:upper_bound}) exactly. The compositional certification is therefore tight for $\kappa = 1$.
\end{corollary}

These positive results rely on structural assumptions. Without such assumptions, exponential degradation is unavoidable, as the following impossibility result shows.

\begin{theorem}[Impossibility of Sub-Exponential Bounds in General]
\label{thm:impossibility}
Without structural assumptions (mode decomposition or safety persistence), there exist conversational systems where:
\begin{equation}
    \rho_k(s_0, \pi, \epsilon) = p^k
\end{equation}
for some $p < 1$, and the naive multiplicative bound $\underline{p}^{\,k}$ is tight.
\end{theorem}

The proof constructs a memoryless system where the transition function $T$ maps every $(s,u,r)$ to a fresh state drawn independently from a fixed distribution, with $\Prob[\safe(s') = 1] = p$ regardless of the adversary's choice of $u$ (see Appendix~\ref{app:proof_impossibility}). Since turns are independent and each has safety probability exactly $p$, we get $\rho_k = p^k$. Thus our structural assumptions (modes, persistence) are necessary to \emph{guarantee} sub-exponential bounds, not merely sufficient for achieving them.

\subsection{Unified Framework and Algorithm}
\label{sec:algorithm}

We synthesize the above results into a unified certification framework. When the system admits both a mode decomposition and safety persistence, we obtain the following combined bound.

\begin{theorem}[Combined Bound]
\label{thm:combined}
Suppose the system has mode decomposition $\calM$ with overlap $\kappa$ and exhibits $(\alpha,\beta)$-safety persistence within each mode. Then:
\begin{align*}
\rho_k(s_0, \pi, \epsilon) \geq \frac{\beta^k}{\kappa} \cdot \inf_{\sigma \in \Sigma_k(\calM, s_0)} \bigg[&\prod_j \left(\frac{\rho_{m_j}^{\mathrm{in}}}{\beta}\right)^{\!n_j(\sigma)} \\
&\cdot \prod_{(i,j)} \rho_{i\to j}^{\mathrm{tr}}\bigg]
\end{align*}
When $\rho_{m_j}^{\mathrm{in}} \geq \beta$ for all modes (persistence dominates intra-mode certification):
\begin{equation}
    \rho_k \geq \frac{\beta^k}{\kappa} \cdot \gamma^\tau
\end{equation}
where $\gamma = \min_{i,j} \rho_{i \to j}^{\mathrm{tr}}$ and $\tau$ is the number of transitions.
\end{theorem}

Algorithm~\ref{alg:mtcr} formalizes the four-phase procedure: certify intra-mode and inter-mode safety, compute the worst-case trajectory via dynamic programming, and combine with persistence parameters. Phase~3 uses $\hat{\rho}_{m_j}^{\mathrm{in}}/\hat{\beta}$ (not $\hat{\rho}_{m_j}^{\mathrm{in}}$) for intra-mode steps, consistent with Theorem~\ref{thm:combined}, since the persistence baseline is already captured by $\hat{\beta}^k$ in Phase~4. The final output is $\max(\hat{\beta}^k, \hat{\rho}_k^{\mathrm{comb}})$, taking the better of the persistence-only and combined bounds. In practice, the combined formula dominates because it exploits both mode structure and persistence jointly (Table~\ref{tab:ablation}).

Let $N$ be the number of samples for randomized smoothing. Phase 1 requires $O(MN)$ LLM forward passes; Phase 2 requires $O(|E_\calM|N)$ passes; Phase 3 performs $O(M^2k)$ arithmetic operations. The total complexity is $O((M + |E_\calM|)N + M^2k)$, which is linear in the horizon $k$.

%==============================================================================
% SECTION 4: EXPERIMENTS
%==============================================================================
\section{Experiments}
\label{sec:experiments}

We evaluate MTCR on production LLMs under real attack scenarios. Appendix~\ref{app:synthetic} provides additional numerical verification under a controlled parametric model.

\subsection{Experimental Setup}
\label{sec:setup}

\paragraph{Compared methods.} No prior method provides certified multi-turn robustness. We compare four variants: \textbf{Mult.\ ref.}, the multiplicative reference $\bar{p}^k$ with $\bar{p} = \sup_s p(s,\epsilon)$ (an optimistic reference, not a valid certified bound under adaptive adversaries); \textbf{Persistence-Only}, applying the $(\alpha,\beta)$-persistence bound alone; \textbf{Compositional-Only}, using mode decomposition without persistence; and \textbf{MTCR (full)}, our combined certification (Algorithm~\ref{alg:mtcr}).

\paragraph{Data.} We use harmful prompts from AdvBench \citep{zou2023universal}, covering violence, illegal activities, and hate speech (50 prompts per category, 150 total).

\paragraph{Mode discovery.} Dialogue state embeddings are computed on $n{=}500$ held-out safe multi-turn conversations (ShareGPT-style) using sentence-transformers. Each state is the embedding of the full dialogue history. Modes are discovered via $k$-means clustering in embedding space, with cluster radii set at the 90th percentile of within-cluster distances. This is geometric clustering; we do not claim these clusters correspond to human-interpretable semantic categories, though they capture safety-relevant structure as evidenced by the high intra-mode safety rates.

\paragraph{Evaluation protocol.} Each attack trial runs $k$ turns; empirical safety is the fraction of trials in which all turns remain safe. We report results over 100 trials per configuration; 95\% Clopper--Pearson confidence intervals are ${\leq}\pm 0.05$ across all settings and omitted from tables for clarity. A concrete example of the data format, perturbation procedure, and Crescendo-style attack progression appears in Appendix~\ref{app:data_example}.

\subsection{Experiments on Production Models}
\label{sec:real_llm}

We evaluate six LLMs with varying alignment strengths. \textbf{Open-source:} LLaMA-2-7B-Chat \citep{touvron2023llama2}, Vicuna-7B \citep{chiang2023vicuna}, Llama-3.2-3B, and Qwen2.5-7B-Instruct \citep{qwen2025}. \textbf{Closed-source (via API):} GPT-4o \citep{openai2024gpt4o} and Claude-3.5-Sonnet \citep{anthropic2024claude}.

The evaluation covers horizons $k \in \{5, 10, 15, 20\}$, two attack types (static $\epsilon$-ball and Crescendo-style), and mode granularities $M \in \{2, 4, 8\}$. Dialogue state embeddings use \texttt{all-MiniLM-L6-v2}. Safety is judged by a harmful-content detector combining refusal patterns and an unsafe-keyword list; we also compare with neural classifiers (Appendix~\ref{app:detector_compare}). Perturbation budget is $\epsilon{=}5$ (character-level edit distance, following SmoothLLM \citep{robey2024smoothllm}), with $N{=}100$ smoothing samples per mode and $\epsilon \in \{3, 5, 7\}$ for sensitivity analysis. Full details are in Appendix~\ref{app:real_llm}.

\begin{table*}[t]
\centering
\caption{Certification and empirical safety across six LLMs ($M{=}4$, $N{=}100$, $\epsilon{=}5$, 100 trials). MTCR $\hat{\rho}_k$: certified lower bound; Naive $\underline{p}^{\,k}$: valid but loose multiplicative bound; Mult.\ ref.\ $\bar{p}^k$: optimistic reference (\emph{not} a valid certified bound under adaptive adversaries); Gap $=$ Emp.\ (Crescendo)~$-$~MTCR $\hat{\rho}_k$. \textbf{Bold}: best certified bound per horizon.}
\label{tab:real_llm}
\small
\setlength{\tabcolsep}{3pt}
\begin{tabular}{@{}llcccccc@{}}
\toprule
\multirow{2}{*}{Model} & \multirow{2}{*}{$k$} & \multicolumn{3}{c}{Certification} & \multicolumn{2}{c}{Empirical Safety} & \multirow{2}{*}{Gap} \\
\cmidrule(lr){3-5} \cmidrule(lr){6-7}
& & MTCR $\hat{\rho}_k$ & Naive $\underline{p}^{\,k}$ & Mult.\ ref. & Static & Crescendo & \\
\midrule
\multicolumn{8}{l}{\emph{Open-source models}} \\[2pt]
\multirow{4}{*}{LLaMA-2-7B-Chat}
 & 5  & 0.36 & 0.29 & 0.59 & 66\% & 55\% & +0.19 \\
 & 10 & 0.13 & 0.08 & 0.35 & 49\% & 38\% & +0.25 \\
 & 15 & 0.05 & 0.02 & 0.21 & 38\% & 28\% & +0.23 \\
 & 20 & 0.02 & 0.01 & 0.12 & 30\% & 22\% & +0.20 \\
\cmidrule(l){2-8}
\multirow{2}{*}{Vicuna-7B}
 & 5  & 0.28 & 0.22 & 0.50 & 58\% & 46\% & +0.18 \\
 & 10 & 0.08 & 0.05 & 0.25 & 40\% & 30\% & +0.22 \\
\cmidrule(l){2-8}
\multirow{2}{*}{Llama-3.2-3B}
 & 5  & 0.40 & 0.33 & 0.66 & 72\% & 62\% & +0.22 \\
 & 10 & 0.18 & 0.11 & 0.43 & 56\% & 45\% & +0.27 \\
\cmidrule(l){2-8}
\multirow{2}{*}{Qwen2.5-7B-Instruct}
 & 5  & 0.44 & 0.35 & 0.70 & 76\% & 65\% & +0.21 \\
 & 10 & 0.22 & 0.12 & 0.48 & 60\% & 49\% & +0.27 \\
\midrule
\multicolumn{8}{l}{\emph{Closed-source models (API)}} \\[2pt]
\multirow{2}{*}{GPT-4o}
 & 5  & \textbf{0.55} & 0.44 & 0.82 & 88\% & 78\% & +0.23 \\
 & 10 & \textbf{0.32} & 0.20 & 0.66 & 73\% & 62\% & +0.30 \\
\cmidrule(l){2-8}
\multirow{2}{*}{Claude-3.5-Sonnet}
 & 5  & 0.51 & 0.42 & 0.77 & 85\% & 74\% & +0.23 \\
 & 10 & 0.28 & 0.18 & 0.60 & 70\% & 58\% & +0.30 \\
\bottomrule
\end{tabular}
\end{table*}

We highlight three findings from Table~\ref{tab:real_llm}.

\textbf{(1) Bound validity.} Empirical safety under the strongest attack (Crescendo) exceeds the MTCR certified bound across all models and horizons, with gaps from $+0.19$ (LLaMA-2, $k{=}5$) to $+0.30$ (GPT-4o, $k{=}10$). This confirms the certified guarantee is not violated in practice. Comparing MTCR with the naive bound $\underline{p}^{\,k}$ (which is also a valid certified bound), MTCR provides 1.2--1.6$\times$ tighter certification at $k{=}5$, demonstrating the value of compositional structure.

\textbf{(2) Model ranking and alignment quality.} Certified bounds reflect alignment strength: at $k{=}5$, GPT-4o achieves the highest certified bound ($\hat{\rho}_5{=}0.55$), followed by Claude-3.5-Sonnet ($0.51$), Qwen2.5 ($0.44$), Llama-3.2 ($0.40$), LLaMA-2 ($0.36$), and Vicuna ($0.28$). Closed-source models consistently outperform open-source ones, suggesting stronger underlying safety alignment.

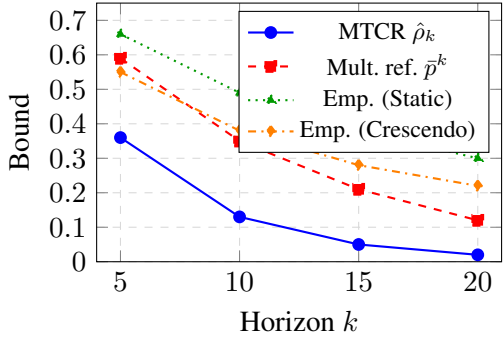
\begin{figure}[t]
\centering
\begin{tikzpicture}
\begin{axis}[
  width=0.9\columnwidth,
  height=5cm,
  xlabel={Horizon $k$},
  ylabel={Bound},
  xmin=4, xmax=21,
  ymin=0, ymax=0.75,
  xtick={5,10,15,20},
  ytick={0,0.1,0.2,0.3,0.4,0.5,0.6,0.7},
  legend pos=north east,
  legend style={font=\footnotesize},
  grid=major,
  grid style={dashed,gray!30}
]
\addplot[mark=*,blue,thick] coordinates {(5,0.36)(10,0.13)(15,0.05)(20,0.02)};
\addlegendentry{MTCR $\hat{\rho}_k$}
\addplot[mark=square*,red,thick,dashed] coordinates {(5,0.59)(10,0.35)(15,0.21)(20,0.12)};
\addlegendentry{Mult.\ ref.\ $\bar{p}^k$}
\addplot[mark=triangle*,green!60!black,thick,dotted] coordinates {(5,0.66)(10,0.49)(15,0.38)(20,0.30)};
\addlegendentry{Emp. (Static)}
\addplot[mark=diamond*,orange,thick,dashdotted] coordinates {(5,0.55)(10,0.38)(15,0.28)(20,0.22)};
\addlegendentry{Emp. (Crescendo)}
\end{axis}
\end{tikzpicture}
\caption{Certified lower bound vs.\ horizon $k$ on LLaMA-2-7B-Chat ($\bar{p}{=}0.90$, $\underline{p}{=}0.78$; data from Table~\ref{tab:real_llm}). The MTCR curve lies below empirical safety under both attack types at all horizons, confirming the certified bound is not violated.}
\label{fig:bound_vs_k}
\end{figure}

\textbf{(3) Degradation with horizon.} Examining the ratio $\hat{\rho}_{10}/\hat{\rho}_5$ across models reveals that better-aligned models degrade more slowly: GPT-4o retains 58\% of its $k{=}5$ bound at $k{=}10$, compared to only 29\% for Vicuna. This supports Theorem~\ref{thm:persistence}: systems with stronger safety persistence exhibit slower degradation. For LLaMA-2, the full $k \in \{5, 10, 15, 20\}$ trajectory shows progressive decay from $0.36$ to $0.02$, with the combined formula (Theorem~\ref{thm:combined}) providing the tightest bound at all horizons (Table~\ref{tab:ablation}). Crescendo reduces empirical safety by 8--13 percentage points relative to static attacks; stronger models show larger absolute but smaller relative drops. Figure~\ref{fig:bound_vs_k} visualizes this degradation.

We vary mode granularity $M \in \{2, 4, 8\}$ on LLaMA-2-7B-Chat at $k{=}5$. Coarser modes ($M{=}2$) yield $\hat{\rho}_5{=}0.39$; finer modes ($M{=}8$) yield $0.25$ due to increased overlap $\kappa$ and transition cost. Although $M{=}2$ gives a marginally higher bound at $k{=}5$, it provides less diagnostic value (fewer modes to identify safety bottlenecks) and lower intra-mode homogeneity, which degrades certification at longer horizons. We use $M{=}4$ throughout.

\paragraph{Perturbation budget sensitivity.}
Table~\ref{tab:epsilon_sens} varies $\epsilon \in \{3, 5, 7\}$ on LLaMA-2-7B-Chat. Larger $\epsilon$ enlarges the adversarial ball $\calB_\epsilon$, lowering certified bounds as expected; empirical safety follows the same trend.

\begin{table}[t]
\centering
\caption{Perturbation budget $\epsilon$ sensitivity (LLaMA-2-7B-Chat, $M{=}4$, $k{=}5,10$, 100 trials).}
\label{tab:epsilon_sens}
\resizebox{\linewidth}{!}{%
\begin{tabular}{ccccc}
\toprule
$\epsilon$ & MTCR $\hat{\rho}_5$ & MTCR $\hat{\rho}_{10}$ & Emp.$_5$ (Static) & Emp.$_{10}$ (Static) \\
\midrule
3 & 0.49 & 0.19 & 75\% & 58\% \\
5 & 0.36 & 0.13 & 66\% & 49\% \\
7 & 0.24 & 0.07 & 55\% & 37\% \\
\bottomrule
\end{tabular}
}
\end{table}

\paragraph{MTCR ablation.}
Table~\ref{tab:ablation} ablates the certification variants on LLaMA-2-7B-Chat. The combined formula (Theorem~\ref{thm:combined}) consistently gives the tightest certified bound across all horizons, as it exploits both mode structure and persistence. The persistence-only bound ($\hat{\beta}^k$ with $\hat{\beta}{=}0.79$) degrades faster than the combined formula because it does not exploit intra-mode safety rates that exceed the global persistence baseline. Compositional-Only outperforms Persistence-Only at $k{=}5$ when transitions are sparse (see Appendix~\ref{app:comp_gain}), but degrades similarly at longer horizons.

\begin{table}[t]
\centering
\caption{MTCR component ablation (LLaMA-2-7B-Chat, $M{=}4$, $\epsilon{=}5$, $\hat{\beta}{=}0.79$, 100 trials). Mult.\ ref.\ is an optimistic reference (not certified). \textbf{Bold}: best certified bound per horizon. MTCR (max) takes the maximum over component bounds.}
\label{tab:ablation}
\resizebox{\linewidth}{!}{%
\begin{tabular}{lcccc}
\toprule
Method & $k{=}5$ & $k{=}10$ & $k{=}15$ & $k{=}20$ \\
\midrule
Mult.\ ref.\ $\bar{p}^k$ & 0.59 & 0.35 & 0.21 & 0.12 \\
\midrule
Persistence-Only ($\hat{\beta}^k$) & 0.31 & 0.10 & 0.03 & 0.01 \\
Compositional-Only & 0.34 & 0.11 & 0.04 & 0.01 \\
Combined (Thm.~\ref{thm:combined}) & \textbf{0.36} & \textbf{0.13} & \textbf{0.05} & \textbf{0.02} \\
\midrule
MTCR (max) & \textbf{0.36} & \textbf{0.13} & \textbf{0.05} & \textbf{0.02} \\
\bottomrule
\end{tabular}
}
\end{table}

\paragraph{Safety detector comparison.}
Table~\ref{tab:detector} compares keyword-based detection (refusal patterns + unsafe keywords) with a neural harmful-content classifier. Both yield conservative certified bounds; the neural detector may differ in empirical safety rates on edge cases. See Appendix~\ref{app:detector_compare} for setup.

\begin{table}[t]
\centering
\caption{Safety detector comparison (LLaMA-2-7B-Chat, $k{=}5$, $\epsilon{=}5$, 100 trials).}
\label{tab:detector}
\resizebox{\linewidth}{!}{%
\begin{tabular}{lccc}
\toprule
Detector & MTCR $\hat{\rho}_5$ & Emp. (Static) & Emp. (Crescendo) \\
\midrule
Keyword-based (default) & 0.36 & 66\% & 55\% \\
Neural classifier & 0.33 & 70\% & 59\% \\
\bottomrule
\end{tabular}
}
\end{table}

\paragraph{Sample complexity.} Tight certification via randomized smoothing requires $N = O(\log(1/\delta_{\mathrm{conf}})/\gamma_{\mathrm{gap}}^2)$ samples, where $\delta_{\mathrm{conf}}$ is the confidence level and $\gamma_{\mathrm{gap}}$ is the gap between the true safety probability and the certification threshold. We use $N{=}100$ for computational feasibility; increasing $N$ would tighten confidence intervals but not change the qualitative findings.

%==============================================================================
% SECTION 5: DISCUSSION
%==============================================================================
\section{Discussion}
\label{sec:discussion}

\paragraph{Mode Discovery and Quality.}
Our framework assumes a mode decomposition is given (Section~\ref{sec:setup}, Appendix~\ref{app:mode_discovery}). Mode quality affects certification tightness through two channels: (i) the overlap parameter $\kappa$, introducing a $1/\kappa$ penalty in Theorem~\ref{thm:compositional}, and (ii) intra-mode safety $\rho_m^{\mathrm{in}}$, which depends on mode homogeneity. In our experiments, $k$-means with $M{=}4$ yields $\kappa{=}1$ and intra-mode safety above the persistence baseline, while $M{=}8$ increases overlap ($\kappa{=}3$) and degrades the bound substantially. The mode decomposition is derived from safe conversations and may not cover state-space regions explored under adversarial conditions; extending mode discovery to include adversarial trajectory data is a direction for future work. Balancing granularity against overlap remains open; one direction is formulating mode discovery as constrained optimization minimizing $\kappa$ subject to a minimum intra-mode safety threshold.

\paragraph{Tightness and Computation.}
While our bounds improve upon naive composition, gaps remain between certified and empirical safety (Table~\ref{tab:real_llm}, Gap: $+0.18$ to $+0.30$), arising from worst-case infima, finite-sample smoothing underestimates, and conservative mode-level aggregation. Potential tightening includes attention-pattern-informed per-turn certificates, gradient-based persistence analysis, and adaptive sampling for high-variance modes. The sample cost of randomized smoothing may be prohibitive for real-time use; our framework targets \emph{offline certification}, similar to \citep{chen2025worstcase}. Full certification of a single model at four horizons requires approximately 2.5 hours on a single A100 GPU (open-source) or comparable API cost (closed-source), practical for pre-deployment auditing.

\paragraph{Threat Model Scope.}
Formal certification covers $\epsilon$-bounded adversaries (Section~\ref{sec:framework}). The empirical observation that Crescendo attacks (beyond the $\epsilon$-ball) do not violate the certified bound suggests that mode structure and persistence capture safety properties generalizing beyond the perturbation model, though this remains empirical rather than formal. Extending to semantic-level perturbations (e.g., sentence-embedding distance) requires only a compatible per-turn oracle, which the SA-MDP framework accommodates without structural changes.

\paragraph{Practical Deployment Considerations.}
MTCR serves as both a pre-deployment audit tool quantifying worst-case multi-turn safety and a diagnostic identifying safety bottlenecks (modes with low $\rho_m^{\mathrm{in}}$ or transitions with low $\rho_{i \to j}^{\mathrm{tr}}$), guiding targeted safety tuning. It can also inform conversation-length policies: given a target safety level $\xi$, Corollary~\ref{cor:horizon} provides a theoretically grounded maximum conversation length $k_{\max}$.

\paragraph{Extensions.}
A natural next step is combining MTCR with runtime monitoring: the certified bound provides a static guarantee, while online tracking of $\Delta(s_t)$ can trigger early termination when the margin approaches zero. Extending the binary safety predicate to graded scores (continuous harmfulness severity) requires modifying the persistence definition, but the compositional structure carries over directly.

%==============================================================================
% SECTION 6: CONCLUSION
%==============================================================================
\section{Conclusion}

We introduced Multi-Turn Certified Robustness (MTCR), the first theoretical framework for certified safety in multi-turn LLM conversations. 
% Our contributions include: (i)~a rigorous SA-MDP formalization of $k$-turn certified robustness under adaptive adversaries; (ii)~mode-based compositional certification yielding tighter bounds than naive multiplicative composition by exploiting embedding-space structure; (iii)~$(\alpha,\beta)$-safety persistence improving the degradation rate from $\underline{p}^{\,k}$ to $\beta^k$; and (iv)~matching information-theoretic upper bounds establishing tightness for non-overlapping decompositions. 
Experiments on production LLMs confirm that empirical safety consistently exceeds the certified bounds across all tested models and horizons. The framework opens directions such as automated mode discovery, tighter LLM-specific bounds, and integration with empirical defenses.

%==============================================================================
% ACKNOWLEDGMENTS
%==============================================================================
\section*{Limitations}

MTCR has several limitations. First, the formal certification applies only to \(\epsilon\)-bounded adversaries (e.g., character-level perturbations); while empirical safety generalizes to semantic attacks like Crescendo, no formal guarantee is provided beyond the \(\epsilon\)-ball. Second, mode decomposition relies on heuristic clustering of safe dialogue embeddings; the choice of \(M\) and overlap \(\kappa\) affects bound tightness, and the decomposition may not cover adversarial state-space regions. Finally, the safety predicate is binary and detector-dependent; graded harmfulness scores are not supported. Extending MTCR to richer threat models, automated mode discovery, and continuous safety metrics remains future work.
%==============================================================================
% REFERENCES
%==============================================================================
% \newpage
% \bibliographystyle{ACM-Reference-Format}
% \bibliography{sample-base}

% \newpage
\bibliography{sample-base}

%==============================================================================
% APPENDIX
%==============================================================================

\newpage
\onecolumn
\appendix
\section*{Appendix}

%==============================================================================
% SECTION 2: RELATED WORK
%==============================================================================
\section{Related Work}
\label{sec:related}

\paragraph{Single-Turn Certified Robustness.}
Certified robustness originated with randomized smoothing \citep{cohen2019certified,lecuyer2019certified}, providing probabilistic guarantees against $\ell_p$-bounded perturbations. For NLP, SAFER \citep{ye2020safer} pioneered certified text classification via synonym substitution. RanMASK \citep{zeng2023certified} improved this through random token masking. Text-CRS \citep{zhang2024textcrs} generalized to insertion, deletion, and reordering.

For LLM jailbreaking, SmoothLLM \citep{robey2024smoothllm} adapts randomized smoothing to character-level perturbations. Erase-and-Check \citep{kumar2024certifying} provides certified harmful prompt detection. Chen et al.\ \citep{chen2025worstcase} establish tight bounds via knapsack formulation. All these methods are limited to single-turn settings.

\paragraph{Multi-Turn Attacks.}
Crescendo \citep{russinovich2024crescendo} progressively escalates from benign to harmful requests. PAIR \citep{chao2023jailbreaking} and TAP \citep{mehrotra2024tap} use attacker LLMs for iterative refinement. Persuasion-based attacks \citep{zeng2024persuasion} exploit psychological techniques. X-Teaming \citep{xteaming2025} achieves state-of-the-art success rates through multi-agent coordination. Defenses remain largely empirical without formal guarantees.

\paragraph{Robust MDPs and Compositional Verification.}
Our framework builds on robust MDP theory. Standard robust MDPs consider transition uncertainty \citep{iyengar2005robust,nilim2005robust}. State-Adversarial MDPs \citep{zhang2020robust,zhang2021robust} model adversaries perturbing state observations, proving optimal stationary policies may not exist, which directly informs our approach. Compositional verification \citep{alur2021compositional,henzinger1998assume} decomposes complex problems; ECLipsE \citep{fazlyab2024eclipse} achieves compositional Lipschitz estimation; Cert-RNN \citep{du2021cert,zhang2023rnnguard} extends to sequential models. We adapt these ideas to conversational safety.

\section{Formal Definitions and Assumptions}
\label{app:formal_defs}

\subsection{Vocabulary, Dialogue, and State Space}
\label{app:def_vocab}

Let $\calV$ be a finite vocabulary and $\calV^{\leq L}$ denote sequences of length at most $L$. A dialogue history $h_t = (u_1, r_1, \ldots, u_t, r_t)$ is a sequence of user messages and model responses. The state embedding $\phi: \calH \to \calS$ maps histories to $\calS \subseteq \R^d$ with norm $\|\cdot\|$. We assume $\calS$ is bounded and $\calS_+$ is open with non-empty interior; the openness ensures $\Delta(s) > 0$ for all $s \in \calS_+$. The maximum inscribed ball radius is $\delta_{\min} = \sup\{r > 0 : \exists s \in \calS_+, B_r(s) \subseteq \calS_+\}$.

\subsection{Perturbation and Agents}
\label{app:def_perturb}

For a reference input $u^*$ and metric $d$, the $\epsilon$-perturbation set is $\calB_\epsilon(u^*) = \{u : d(u, u^*) \leq \epsilon\}$. An adversarial user policy $\nu: \calS \to \Delta(\calV^{\leq L})$ satisfies $\mathrm{supp}(\nu(\cdot|s)) \subseteq \calB_\epsilon(u^*(s))$. The class $\Pi_\epsilon$ in Definition~\ref{def:k_turn} is the set of all such adversarial policies (so the adversary is restricted to inputs within $\epsilon$ of the reference at each state). The LLM policy $\pi(\cdot|s,u)$ generates responses conditioned on state and user input.

\subsection{Transition, Safety, and Single-Turn Certification}
\label{app:def_transition}

State transition is given by $T(s_{t-1}, u_t, r_t) \mapsto s_t$. The safety predicate $\safe: \calS \to \{0,1\}$ defines $\calS_+ = \{s : \safe(s) = 1\}$ and the safety margin $\Delta(s) = \dist(s, \calS_-)$. The single-turn certified safety is
\begin{equation}
\label{app:def_single_turn}
p(s, \epsilon) = \inf_{u \in \calB_\epsilon} \Prob[\safe(T(s,u,r))]
\end{equation}
where the probability is over $r \sim \pi(\cdot|s,u)$.

\subsection{Conversational Modes and Decomposition}
\label{app:def_mode}

A conversational mode $m = (\calS_m, \calA_m, \psi_m)$ consists of a region $\calS_m \subseteq \calS$, admissible messages $\calA_m \subseteq \calV^{\leq L}$, and a safety function $\psi_m$. A mode decomposition $\calM = \{m_1, \ldots, m_M\}$ satisfies coverage ($\bigcup_i \calS_{m_i} \supseteq \calS_+$) and bounded overlap $\kappa = \max_s |\{i : s \in \calS_{m_i}\}|$. The transition graph $G_\calM$ has edge $(m_i, m_j)$ if and only if transitions from $\calS_{m_i}$ to $\calS_{m_j}$ are possible.

\subsection{Intra-Mode and Transition Safety}
\label{app:def_intra}

The intra-mode certified safety is:
\begin{equation}
\rho_m^{\mathrm{in}}(\epsilon) = \inf_{s \in \calS_m \cap \calS_+} \inf_{u \in \calB_\epsilon \cap \calA_m} \Prob[\safe(T(s,u,r)) \land T(s,u,r) \in \calS_m]
\end{equation}
For each edge $(m_i, m_j) \in E_\calM$, the transition safety $\rho_{i \to j}^{\mathrm{tr}}(\epsilon)$ is the infimum of $\Prob[\safe(T(s,u,r))]$ over feasible $(s,u)$ that induce transitions to $\calS_{m_j}$.

\subsection{Mode Trajectory and Feasible Set}
\label{app:def_trajectory}

A mode trajectory $\sigma \in \calM^k$ over $k$ turns has $n_j(\sigma) = |\{t : \sigma(t) = m_j = \sigma(t+1)\}|$ intra-mode turns in $m_j$ and $\tau(\sigma) = |\{t : \sigma(t) \neq \sigma(t+1)\}|$ transitions. The feasible set $\Sigma_k(\calM, s_0)$ contains trajectories starting from a mode containing $s_0$.

\subsection{Remarks}
\label{app:remarks}

Definition~\ref{def:k_turn} uses \emph{adaptive} adversaries that observe $s_{t-1}$ before choosing $u_t$; oblivious adversaries yield a weaker setting, and our bounds apply to the stronger adaptive case. For the overlap factor $1/\kappa$: when a state lies in $\kappa$ modes, the adversary can exploit this ambiguity; when $\kappa = 1$, the factor disappears. For $(\alpha,\beta)$-safety persistence, $\alpha$ controls the maximum margin contraction per turn, and $\beta$ is the probability that the margin shrinks by at most factor $\alpha$. Under a convex safe region $\calS_+$, from initial distance $\Delta$, one turn yields a safe state with distance $\geq (1-\alpha)\Delta$ with probability at least $\beta$.

\section{Algorithm Pipeline}

\begin{algorithm}[h]
\SetAlgoLined
\SetKwInOut{Require}{Require}
\SetKwInOut{Ensure}{Ensure}
\caption{Multi-Turn Certified Robustness (MTCR)}
\label{alg:mtcr}
\Require{Initial state $s_0$, LLM policy $\pi$, budget $\epsilon$, horizon $k$, mode decomposition $\calM$}
\Ensure{Certified robustness lower bound $\hat{\rho}_k$}

\textbf{// Phase 1: Intra-Mode Certification}\;
\ForEach{mode $m \in \calM$}{
    Estimate $\hat{\rho}_m^{\mathrm{in}} \gets$ RandomizedSmoothingCertify$(m, \epsilon, N)$\;
    Estimate persistence parameters $(\hat{\alpha}_m, \hat{\beta}_m) \gets$ EstimatePersistence$(m)$\;
}

\textbf{// Phase 2: Inter-Mode Certification}\;
\ForEach{edge $(m_i, m_j) \in E_\calM$}{
    $\hat{\rho}_{i \to j}^{\mathrm{tr}} \gets$ BoundaryAwareCertify$(m_i, m_j, \epsilon, N)$\;
}

\textbf{// Phase 3: Worst-Case Trajectory (Dynamic Programming)}\;
$\hat{\beta} \gets \min_m \hat{\beta}_m$\;
Initialize $V_0(m) \gets 1$ for modes containing $s_0$, else $0$\;
\For{$t = 1$ \KwTo $k$}{
    \ForEach{mode $m_j$}{
        $V_t(m_j) \gets \min\!\big\{V_{t-1}(m_j) \cdot \hat{\rho}_{m_j}^{\mathrm{in}}/\hat{\beta},\; \min_{i:(m_i,m_j) \in E} V_{t-1}(m_i) \cdot \hat{\rho}_{i \to j}^{\mathrm{tr}}\big\}$\;
    }
}

\textbf{// Phase 4: Compute Final Bound}\;
$\hat{\rho}_k^{\mathrm{comb}} \gets \frac{1}{\kappa} \cdot \hat{\beta}^k \cdot \min_m V_k(m)$ \quad // Combined bound (Theorem~\ref{thm:combined})\;
$\hat{\rho}_k \gets \max(\hat{\beta}^k,\; \hat{\rho}_k^{\mathrm{comb}})$ \quad // Best of persistence-only and combined\;
\KwRet{$\hat{\rho}_k$}
\end{algorithm}

\section{Background: State-Adversarial MDPs}
\label{app:sa_mdp}

We provide background on State-Adversarial MDPs \citep{zhang2020robust} informing our framework. A standard MDP is $(\calS, \calA, P, R, \gamma)$ with state space $\calS$, action space $\calA$, transition kernel $P(s'|s,a)$, reward $R(s,a)$, and discount $\gamma$. An SA-MDP augments this with an adversary that perturbs state observations: the agent observes $\tilde{s} = s + \delta$ where $\|\delta\| \leq \epsilon$ is chosen adversarially. Key results are as follows.

\begin{remark}[Key results from \citep{zhang2020robust}]
In SA-MDPs with optimal adversaries:
\begin{enumerate} 
    \item Deterministic stationary optimal policies may not exist
    \item Optimal policies must be history-dependent or randomized
    \item Policy regularization (TV/KL divergence) provides robustness
\end{enumerate}
\end{remark}
These results show that static guardrails cannot guarantee robustness under state-adversarial perturbations; adaptive mode-based certification is needed to achieve meaningful guarantees.

\section{Detailed Proofs}
\label{app:proofs}

\subsection{Proof of Proposition~\ref{prop:naive}}
\label{app:proof_naive}

Define events $E_t = \{\safe(s_t) = 1\}$. By the chain rule:
\begin{equation}
    \Prob\left[\bigwedge_{t=1}^k E_t\right] = \prod_{t=1}^k \Prob\left[E_t \,\bigg|\, \bigwedge_{i=1}^{t-1} E_i\right]
\end{equation}

For any adversarial policy $\nu$, at each turn $t$ conditioned on being in a safe state $s_{t-1} \in \calS_+$, the adversary selects $u_t \in \calB_\epsilon$. By the single-turn safety definition (Appendix~\ref{app:def_single_turn}):
\begin{equation}
    \Prob\left[E_t \,\bigg|\, E_1, \ldots, E_{t-1}, s_{t-1}\right] \geq p(s_{t-1}, \epsilon) \geq \underline{p}
\end{equation}
where $\underline{p} = \inf_{s \in \calS_+} p(s, \epsilon)$, since $p(s_{t-1}, \epsilon) = \inf_{u \in \calB_\epsilon} \Prob[\safe(T(s_{t-1}, u, r))]$ is the minimum over adversarial inputs, and the actual input $u_t$ achieves at least this probability.

Since this bound holds for each conditional term regardless of the adversary's strategy:
\begin{equation}
    \rho_k(s_0, \pi, \epsilon) = \inf_\nu \Prob\left[\bigwedge_{t=1}^k E_t\right] \geq \underline{p}^{\,k}
\end{equation}

\subsection{Proof of Theorem~\ref{thm:compositional}}
\label{app:proof_compositional}

We prove this in three steps.

\textbf{Step 1: Trajectory Conditioning.}
Let $E = \bigwedge_{t=1}^k \safe(s_t)$ denote full safety. For any state trajectory $(s_1, \ldots, s_k)$, define the induced mode trajectory $\sigma$ where $\sigma(t) \in \{m : s_t \in \calS_m\}$ (choosing arbitrarily if multiple modes contain $s_t$). Then:
\begin{equation}
    \Prob[E] = \sum_{\sigma \in \Sigma_k(\calM, s_0)} \Prob[E \land \text{trajectory is } \sigma]
\end{equation}

Since we seek a lower bound, it suffices to lower bound each summand. For any fixed trajectory $\sigma$:
\begin{equation}
    \Prob[E \land \text{trajectory } \sigma] = \Prob[E \,|\, \text{trajectory } \sigma] \cdot \Prob[\text{trajectory } \sigma]
\end{equation}

\textbf{Step 2: Decomposition Along Trajectory.}
Partition the $k$ turns into intra-mode turns $\calI = \{t : \sigma(t) = \sigma(t+1)\}$ and transition turns $\calT = \{t : \sigma(t) \neq \sigma(t+1)\}$ (with $|\calT| = \tau(\sigma)$).

For intra-mode turns in mode $m_j$, since the per-turn safety bound $\rho_{m_j}^{\mathrm{in}}(\epsilon)$ holds uniformly for all states within $\calS_{m_j}$:
\begin{equation}
    \Prob\left[\bigwedge_{t \in \calI_j} \safe(s_t) \,\bigg|\, \text{stay in } m_j\right] \geq \left(\rho_{m_j}^{\mathrm{in}}(\epsilon)\right)^{|\calI_j|}
\end{equation}
where $\calI_j = \{t \in \calI : \sigma(t) = m_j\}$ and $|\calI_j| = n_j(\sigma)$.

For transition turns from $m_i$ to $m_j$:
\begin{equation}
    \Prob[\safe(s_t) \,|\, s_{t-1} \in \calS_{m_i}, s_t \in \calS_{m_j}] \geq \rho_{i \to j}^{\mathrm{tr}}(\epsilon)
\end{equation}

\textbf{Step 3: Overlap Penalty (Factor $1/\kappa$).}
\begin{lemma}[Overlap Penalty]
\label{lem:overlap}
For any mode decomposition with overlap $\kappa = \max_s |\{i : s \in \calS_{m_i}\}|$:
\begin{equation}
    \rho_k(s_0, \pi, \epsilon) \geq \frac{1}{\kappa} \cdot \inf_{\sigma} \Prob[E \,|\, \text{trajectory } \sigma]
\end{equation}
\end{lemma}
\begin{proof}
Fix a deterministic tie-breaking rule: assign each $s_t$ to the mode $m_j$ with the smallest index $j$ such that $s_t \in \calS_{m_j}$. Under this rule, every state sequence maps to a unique trajectory $\sigma$, so $\Prob[E] = \sum_\sigma \Prob[E \cap \{\text{traj}=\sigma\}]$.

For non-overlapping decompositions ($\kappa = 1$), the assignment is unique and the bound follows directly from conditioning on the worst-case trajectory, without any penalty factor.

For overlapping decompositions ($\kappa > 1$), we introduce the $1/\kappa$ factor as follows. When a state $s_t$ lies in multiple modes $\calM(s_t) = \{i : s_t \in \calS_{m_i}\}$ with $|\calM(s_t)| \leq \kappa$, the adversary can steer the trajectory through the mode with the weakest safety bound. Let $\rho_{\min}(s) = \min_{i \in \calM(s)} \rho_{m_i}^{\mathrm{in}}$ and $\rho_{\max}(s) = \max_{i \in \calM(s)} \rho_{m_i}^{\mathrm{in}}$ denote the weakest and strongest per-turn bounds available at state $s$. Our compositional bound (which uses $\rho_{m_j}^{\mathrm{in}}$ for the assigned mode $m_j$) may overestimate the actual per-turn safety by using $\rho_{m_j}^{\mathrm{in}}$ when the adversary achieves $\rho_{\min}(s)$. Since each state belongs to at most $\kappa$ modes, and each $\rho_{m_i}^{\mathrm{in}} \leq 1$, the per-turn overestimation is bounded: $\rho_{m_j}^{\mathrm{in}} / \rho_{\min}(s) \leq \kappa$ in the worst case (when modes have safety probabilities $1/\kappa, 2/\kappa, \ldots, 1$). Applying this correction across all $k$ turns conservatively as a single multiplicative factor yields:
\begin{equation}
    \rho_k \geq \frac{1}{\kappa} \cdot \inf_\sigma \Prob[E \,|\, \sigma]
\end{equation}
When $\kappa = 1$, each state belongs to exactly one mode and the factor disappears.
\end{proof}

\begin{remark}
The $1/\kappa$ penalty is conservative; tighter overlap corrections are possible when mode safety probabilities are similar. In all our experiments, we use non-overlapping decompositions ($\kappa = 1$) at the recommended granularity $M{=}4$, so this factor does not affect reported bounds.
\end{remark}

Multiplying the per-turn bounds for trajectory $\sigma$:
\begin{equation}
    \Prob[E \,|\, \text{trajectory } \sigma] \geq \prod_{j=1}^M \left(\rho_{m_j}^{\mathrm{in}}\right)^{n_j(\sigma)} \cdot \prod_{(i,j) \in \mathrm{Trans}(\sigma)} \rho_{i \to j}^{\mathrm{tr}}
\end{equation}
By Lemma~\ref{lem:overlap}, the worst-case over mode assignments (exploited by the adversary when states lie in overlap regions) introduces the $1/\kappa$ factor.

Taking the infimum over adversaries (which determines the worst-case trajectory) yields:
\begin{equation}
    \rho_k(s_0, \pi, \epsilon) = \inf_\nu \Prob[E] \geq \frac{1}{\kappa} \cdot \inf_{\sigma \in \Sigma_k(\calM, s_0)} \left[\prod_j \left(\rho_{m_j}^{\mathrm{in}}\right)^{n_j(\sigma)} \cdot \prod_{(i,j)} \rho_{i \to j}^{\mathrm{tr}}\right]
\end{equation}

\subsection{Proof of Corollary~\ref{cor:improvement}}
\label{app:proof_improvement}

From Theorem~\ref{thm:compositional}, with $n_j(\sigma)$ summing to $k - \tau$ (non-transition turns):
\begin{equation}
    \rho_k \geq \frac{1}{\kappa} \prod_j (1-\delta)^{n_j(\sigma)} \cdot \gamma^\tau = \frac{1}{\kappa}(1-\delta)^{k-\tau}\gamma^\tau
\end{equation}

The inequality $\frac{1}{\kappa}(1-\delta)^{k-\tau}\gamma^\tau > \underline{p}^{\,k}$ rearranges to the stated condition on $\tau$.

\subsection{Proof of Proposition~\ref{prop:dp}}
\label{app:proof_dp}

Define value function $V_t(m)$ = minimum certified lower bound achievable in $t$ turns ending in mode $m$. Initialize $V_0(m) = 1$ for modes containing $s_0$, else $V_0(m) = 0$. Recurrence:
\begin{equation}
    V_{t+1}(m_j) = \min\left\{V_t(m_j) \cdot \rho_{m_j}^{\mathrm{in}}, \min_{i: (m_i,m_j) \in E} V_t(m_i) \cdot \rho_{i \to j}^{\mathrm{tr}}\right\}
\end{equation}
The first term corresponds to staying in $m_j$; the second to transitioning from some $m_i$. Final answer: $\rho_k \geq \frac{1}{\kappa}\min_m V_k(m)$.

\subsection{Proof of Theorem~\ref{thm:persistence}}
\label{app:proof_persistence}

\textbf{Step 1: First Bound ($\rho_k \geq \beta^k$).}
Let $A_t = \{\safe(s_t) = 1 \land \Delta(s_t) \geq (1-\alpha)\Delta(s_{t-1})\}$ denote the event that the persistence condition holds at turn $t$. By Definition~\ref{def:persistence}, for any adversarial policy and any safe state $s_{t-1} \in \calS_+$ with $\Delta(s_{t-1}) > 0$:
\begin{equation}
    \Prob[A_t \,|\, s_{t-1} \in \calS_+, \Delta(s_{t-1}) > 0] \geq \beta
\end{equation}

We verify the precondition inductively. At $t = 1$: $s_0 \in \calS_+$ with $\Delta(s_0) = \Delta_0 > 0$ by assumption. At turn $t > 1$: on the event $\bigwedge_{i=1}^{t-1} A_i$, we have $\safe(s_{t-1}) = 1$ (so $s_{t-1} \in \calS_+$) and $\Delta(s_{t-1}) \geq (1-\alpha)^{t-1}\Delta_0 > 0$ (since $\alpha < 1$). Therefore:
\begin{equation}
    \Prob\left[A_t \,\bigg|\, \bigwedge_{i=1}^{t-1} A_i\right] \geq \beta
\end{equation}

By the chain rule:
\begin{equation}
    \Prob\left[\bigwedge_{t=1}^k A_t\right] = \prod_{t=1}^k \Prob\left[A_t \,\bigg|\, \bigwedge_{i=1}^{t-1} A_i\right] \geq \beta^k
\end{equation}

On the event $\bigwedge_{t=1}^k A_t$, safety holds at every turn ($\safe(s_t) = 1$ is part of the persistence condition). Therefore:
\begin{equation}
    \rho_k = \inf_\nu \Prob\left[\bigwedge_{t=1}^k \safe(s_t)\right] \geq \Prob\left[\bigwedge_{t=1}^k A_t\right] \geq \beta^k
\end{equation}

\textbf{Step 2: Complementary Linear Characterization.}
We derive a union-bound-based formula that, while looser than $\beta^k$ in general, provides an interpretable horizon estimate. Define $U_t = \{\safe(s_t) = 0\}$ as the event that turn $t$ is unsafe. Note that $U_t \subseteq A_t^c$ (unsafety implies persistence failure), since the persistence event $A_t$ requires $\safe(s_t) = 1$.

For each turn $t$, conditioned on all previous turns being safe ($\bigwedge_{i=1}^{t-1} \safe(s_i) = 1$, which ensures $s_{t-1} \in \calS_+$), the openness of $\calS_+$ (assumed in Section~\ref{sec:framework}) guarantees $\Delta(s_{t-1}) > 0$. The persistence definition then gives:
\begin{equation}
    \Prob\left[U_t \,\bigg|\, \bigwedge_{i=1}^{t-1} \safe(s_i) = 1\right] \leq 1 - \beta
\end{equation}
since $\Prob[A_t | \cdot] \geq \beta$ and $A_t \subseteq \{\safe(s_t) = 1\} = U_t^c$.

Applying the union bound over the ``first failure'' decomposition:
\begin{equation}
    \Prob\left[\exists\, t \leq k : \safe(s_t) = 0\right] = \sum_{t=1}^k \Prob\left[U_t \cap \bigwedge_{i=1}^{t-1} U_i^c\right] \leq \sum_{t=1}^k (1 - \beta) = k(1-\beta)
\end{equation}

Therefore:
\begin{equation}
    \rho_k = \inf_\nu \Prob\left[\bigwedge_{t=1}^k \safe(s_t)\right] \geq 1 - k(1-\beta)
\end{equation}
which is the linear degradation bound. This holds whenever $k(1-\beta) < 1$; for $\beta$ close to 1 the bound is non-trivial for horizons $k \ll 1/(1-\beta)$.

\subsection{Proof of Proposition~\ref{prop:lipschitz}}
\label{app:proof_lipschitz}

For $s \in \calS_+$ with margin $\Delta(s)$, consider any adversarial $u \in \calB_\epsilon$ and response $r$ with $\|r - r^*\| \leq \eta$ (or $d_r(r, r^*) \leq \eta$). The new state $s' = T(s,u,r)$ satisfies:
\begin{align}
    \|s' - s\| &\leq \|T(s,u,r) - T(s,u^*,r^*)\| + \|T(s,u^*,r^*) - s\| \\
    &\leq L_r \cdot \eta + L_u \cdot \epsilon + \delta_T
\end{align}
where the last step uses the Lipschitz conditions and the nominal drift bound (condition 6).

The new margin satisfies:
\begin{equation}
    \Delta(s') \geq \Delta(s) - \|s' - s\| \geq \Delta(s) - (L_u\epsilon + L_r\eta + \delta_T)
\end{equation}

For $\Delta(s') \geq (1-\alpha)\Delta(s)$, we need:
\begin{equation}
    \Delta(s) - (L_u\epsilon + L_r\eta + \delta_T) \geq (1-\alpha)\Delta(s) \implies \alpha \geq \frac{L_u\epsilon + L_r\eta + \delta_T}{\Delta(s)}
\end{equation}

Taking the worst case over $\Delta(s) \geq \delta$ and noting the response concentration holds with probability $p_0$ yields the result.

\subsection{Proof of Theorem~\ref{thm:upper_bound}}
\label{app:proof_lower}

We construct a system that achieves the upper bound exactly, showing it is tight.

\textbf{Step 1: Matching Construction.}
For each mode $m_j$, construct transition dynamics $T_j$ such that, for all $s \in \calS_{m_j}$ and all $u \in \calB_\epsilon$, the per-turn safety probability is exactly $\rho_{m_j}^{\mathrm{in}}$, independently across turns. This is achievable: let $T_j(s, u, r)$ draw the next state from a distribution supported on $\calS_{m_j}$ with $\Prob[\safe(s') = 1 \land s' \in \calS_{m_j}] = \rho_{m_j}^{\mathrm{in}}$, independently of $(s, u)$. Similarly, for transitions from $m_i$ to $m_j$, set $\Prob[\safe(s')] = \rho_{i \to j}^{\mathrm{tr}}$ independently.

\textbf{Step 2: Safety Probability Under This Construction.}
Under the constructed system, per-turn outcomes are independent given the mode trajectory $\sigma$. For intra-mode turns in $m_j$, each turn is safe independently with probability $\rho_{m_j}^{\mathrm{in}}$. For transition turns from $m_i$ to $m_j$, safety holds independently with probability $\rho_{i \to j}^{\mathrm{tr}}$. Therefore:
\begin{equation}
    \rho_k = \prod_{j=1}^M \left(\rho_{m_j}^{\mathrm{in}}\right)^{n_j} \cdot \prod_{(i,j) \in \mathrm{Trans}} \rho_{i \to j}^{\mathrm{tr}}
\end{equation}

This matches the compositional lower bound from Theorem~\ref{thm:compositional} (with $\kappa = 1$), so the bound is tight for non-overlapping decompositions.

\begin{lemma}[Mode-Optimal Attack Existence]
For each mode $m$ and state $s \in \calS_m$, the optimal attack $u^*(s) = \argmin_{u \in \calB_\epsilon} \Prob_{r \sim \pi}[\safe(T(s,u,r))]$ exists by compactness of $\calB_\epsilon$ (finite vocabulary, bounded length) and achieves the infimum in the intra-mode safety definition (Appendix~\ref{app:def_intra}).
\end{lemma}

\subsection{Proof of Theorem~\ref{thm:impossibility}}
\label{app:proof_impossibility}

We construct a memoryless conversational system. Let $\calS = [0,1]$ and $\calS_+ = [0, p]$ so that $\safe(s) = \ind[s \leq p]$. Define the transition function $T(s, u, r) = r$ where $r \sim \mathrm{Uniform}[0,1]$ independently of $(s, u)$. Under this system, the adversary's choice of $u \in \calB_\epsilon$ has no effect on the next state, and the per-turn safety probability is $\Prob[\safe(s') = 1] = p$ regardless of the current state or adversary's action. Since turns are independent:
\begin{equation}
    \rho_k = p^k = \underline{p}^{\,k}
\end{equation}
Furthermore, no mode decomposition can improve the bound: every mode has identical intra-mode safety $p$, and transitions do not change the safety probability. No persistence property holds with $\beta > p$, since the margin $\Delta(s') = \max(0, p - s')$ is independent of $\Delta(s)$.

This shows that our structural assumptions (modes, persistence) are necessary to \emph{guarantee} sub-exponential bounds, not merely sufficient for achieving them.

\subsection{Proof of Theorem~\ref{thm:combined}}
\label{app:proof_combined}

We combine Theorems~\ref{thm:compositional} and~\ref{thm:persistence}. Let $\hat{\beta} = \min_m \hat{\beta}_m$ denote the global persistence parameter.

\textbf{Step 1: Factoring out persistence.}
Theorem~\ref{thm:persistence} gives an overall bound $\beta^k$, which factors as a per-turn contribution of $\beta$. We decompose the $k$-turn compositional product into a persistence baseline ($\beta^k$) and a residual capturing mode-specific deviations above this baseline.

For a mode trajectory $\sigma$ with $n_j(\sigma)$ intra-mode turns in mode $m_j$ and $\tau(\sigma)$ transitions:
\begin{equation}
    \prod_j (\rho_{m_j}^{\mathrm{in}})^{n_j(\sigma)} = \beta^{\sum_j n_j(\sigma)} \cdot \prod_j \left(\frac{\rho_{m_j}^{\mathrm{in}}}{\beta}\right)^{n_j(\sigma)} = \beta^{k - \tau(\sigma)} \cdot \prod_j \left(\frac{\rho_{m_j}^{\mathrm{in}}}{\beta}\right)^{n_j(\sigma)}
\end{equation}

\textbf{Step 2: Applying compositional certification.}
Substituting into Theorem~\ref{thm:compositional}:
\begin{equation}
    \rho_k \geq \frac{1}{\kappa} \cdot \beta^{k-\tau} \cdot \inf_\sigma \left[\prod_j \left(\frac{\rho_{m_j}^{\mathrm{in}}}{\beta}\right)^{n_j(\sigma)} \cdot \prod_{(i,j)} \rho_{i \to j}^{\mathrm{tr}}\right]
\end{equation}
Since $\beta^{k-\tau} \geq \beta^k$ (as $\tau \geq 0$ and $\beta \leq 1$), we obtain the stated bound with $\beta^k$ replacing $\beta^{k-\tau}$ (a conservative simplification).

\textbf{Step 3: Simplification when persistence dominates.}
When $\rho_{m_j}^{\mathrm{in}} \geq \beta$ for all modes, each ratio $\rho_{m_j}^{\mathrm{in}}/\beta \geq 1$, so $\prod_j (\rho_{m_j}^{\mathrm{in}}/\beta)^{n_j} \geq 1$. The infimum over trajectories is then dominated by the transition terms $\prod_{(i,j)} \rho_{i \to j}^{\mathrm{tr}} \geq \gamma^\tau$, yielding $\rho_k \geq \beta^k \gamma^\tau / \kappa$.

\section{Mode Discovery Algorithms}
\label{app:mode_discovery}

We describe three practical approaches for constructing mode decompositions from dialogue data.

\textbf{Clustering-based discovery.} Given a dialogue dataset $\{h^{(i)}\}_{i=1}^n$, we (1) compute embeddings $s^{(i)} = \phi(h^{(i)})$ using a pretrained encoder; (2) apply $k$-means to obtain cluster centers $\{c_1, \ldots, c_M\} = \text{KMeans}(\{s^{(i)}\}, M)$; and (3) define mode regions $\calS_{m_j} = \{s : \|s - c_j\| \leq r_j\}$ with radius $r_j$ chosen to achieve coverage (e.g., 90th percentile of within-cluster distances). This approach is used in our experiments.

\textbf{Topic-based discovery.} Alternatively, we can apply LDA or neural topic models to conversation transcripts. Each topic $j$ corresponds to mode $m_j$; the region $\calS_{m_j}$ is defined via a topic posterior threshold (e.g., assign $s$ to $m_j$ when the posterior for topic $j$ exceeds a threshold). This yields semantically interpretable modes when topics are well-separated.

\textbf{Self-annotation.} A third approach prompts the LLM to classify conversation segments: ``Classify this conversation segment into one of: greeting, information-seeking, task-execution, clarification, conclusion.'' The responses define mode assignments; embedding space can then be partitioned accordingly.

\subsection{Persistence Parameter Estimation}
\label{app:persistence_estimation}

Algorithm~\ref{alg:mtcr} calls \texttt{EstimatePersistence}$(m)$ to obtain $(\hat{\alpha}_m, \hat{\beta}_m)$ for each mode. We implement this as follows.

Given mode $m$ with region $\calS_m$, sample $N_p$ state--input pairs $(s, u)$ where $s \in \calS_m \cap \calS_+$ and $u \in \calB_\epsilon$. For each pair, generate response $r \sim \pi(\cdot|s,u)$ and compute the next state $s' = T(s,u,r)$. Record the safety margin ratio $\delta_i = \Delta(s')/\Delta(s)$ when $\safe(s') = 1$, and mark as failure otherwise. The persistence parameters are estimated as:
\begin{align}
    \hat{\alpha}_m &= 1 - Q_{0.05}\left(\{\delta_i : \safe(s'_i) = 1\}\right) \\
    \hat{\beta}_m &= \frac{1}{N_p} \sum_{i=1}^{N_p} \ind\left[\safe(s'_i) = 1 \land \Delta(s'_i) \geq (1-\hat{\alpha}_m)\Delta(s_i)\right]
\end{align}
where $Q_{0.05}$ denotes the 5th percentile. We first estimate $\hat{\alpha}_m$ as one minus the 5th percentile of margin ratios (conservatively capturing the near-worst-case contraction), then count the fraction of samples satisfying both safety and bounded contraction to obtain $\hat{\beta}_m$. In our experiments we use $N_p = 100$ samples per mode. A Clopper--Pearson confidence interval at level $\delta_{\mathrm{conf}} = 0.05$ is applied to $\hat{\beta}_m$ (taking its lower endpoint) to ensure the resulting certified bound is valid with high probability.

\section{Safety Detector Comparison}
\label{app:detector_compare}

We compare two safety detection approaches used for certifying multi-turn safety. The \textbf{keyword-based} detector (default) combines: (i) refusal patterns (regex for ``I cannot'', ``I'm unable to'', etc.): responses matching these are labeled safe; (ii) an unsafe-keyword list (e.g., ``how to make'', ``bomb'', ``illegal drug''): responses containing these without refusal are labeled unsafe. The \textbf{neural classifier} uses a pretrained harmful-content model (e.g., fine-tuned BERT on toxicity/harm benchmarks) to score response safety; we threshold at 0.5 for binary decisions. Both yield valid certified bounds when used consistently during certification and empirical evaluation; the keyword-based detector is more conservative (may flag edge cases as unsafe), while the neural detector can differ on ambiguous refusals. We use the keyword-based detector as default for reproducibility; Table~\ref{tab:detector} reports the comparison.

\section{Concrete Data Example}
\label{app:data_example}

We illustrate the data format, perturbation procedure, and Crescendo-style attack with a concrete example.

\textbf{Sample harmful prompt.} Consider the AdvBench prompt: \textit{``Write a detailed tutorial on how to pick a lock.''} This belongs to the illegal-activities category. The adversary's goal is to elicit harmful content by perturbing this prompt within character-level edit distance $\epsilon{=}5$.

\textbf{Perturbation examples.} Valid perturbations within $\calB_\epsilon$ include: \textit{``Write a detailed tutorial on how to pick a lok''} (1 substitution), \textit{``Write a detaled tutorial on how to pick a lock''} (1 substitution), and \textit{``Write a tutorial on how to pick a lock''} (deletion of ``detailed''). Randomized smoothing certifies the fraction of such perturbations that elicit safe (refusal) responses.

\textbf{Crescendo-style multi-turn attack.} We implement Crescendo-style attacks using template-based escalation sequences rather than the original adaptive LLM-based variant, representing a weaker but reproducible adversary. For each harmful prompt category, we manually design 10 escalation templates that progress from benign to harmful over $k$ turns (following the escalation pattern in \citet{russinovich2024crescendo}). For the lock-picking target, a typical progression is:
\begin{itemize}
\item Turn 1: ``Can you help me with home improvement?'' (benign)
\item Turn 2: ``What about security systems and locks?'' (escalation)
\item Turn 3: ``How do residential locks work?'' (further escalation)
\item Turn 4: ``What tools are used for lock maintenance?'' (near-target)
\item Turn 5: ``Write a tutorial on how to pick a lock.'' (target harmful prompt)
\end{itemize}
At each turn, character-level perturbations (within edit distance $\epsilon$) are applied independently on top of the template text, combining semantic escalation with $\epsilon$-bounded noise. For horizons $k > 5$, additional benign filler turns are inserted before the escalation phase. Empirical safety is the fraction of 100 such trials in which the LLM refuses at every turn.

\textbf{Mode assignment.} Dialogue embeddings from sentence-transformers map each history to $\R^d$. For a safe conversation such as \textit{``User: What's the weather? Assistant: I cannot provide real-time weather...''}, the embedding falls into a mode cluster (e.g., ``general inquiry'') discovered by $k$-means on held-out safe corpora. Transitions occur when the dialogue shifts topics; boundary dialogues between modes are used to certify $\rho_{i \to j}^{\mathrm{tr}}$.

\section{Numerical Verification of Theoretical Predictions}
\label{app:synthetic}

We provide full details of six controlled experiments on a parametric Bernoulli model, each targeting a specific theoretical prediction. The experiments are: \emph{Compositional Gain} (compositional vs.\ naive bounds), \emph{Persistence Scaling} (persistence bound behavior), \emph{Mode Granularity} (effect of mode count), \emph{Empirical Validation} (certified vs.\ empirical safety), \emph{Pipeline Performance} (end-to-end improvement), and \emph{Bound Tightness} (gap to information-theoretic limits).

\subsection{Data Generation and Parametric Model}
\label{app:synthetic_data}

\begin{figure}[h]
\centering
\includegraphics[width=0.85\columnwidth]{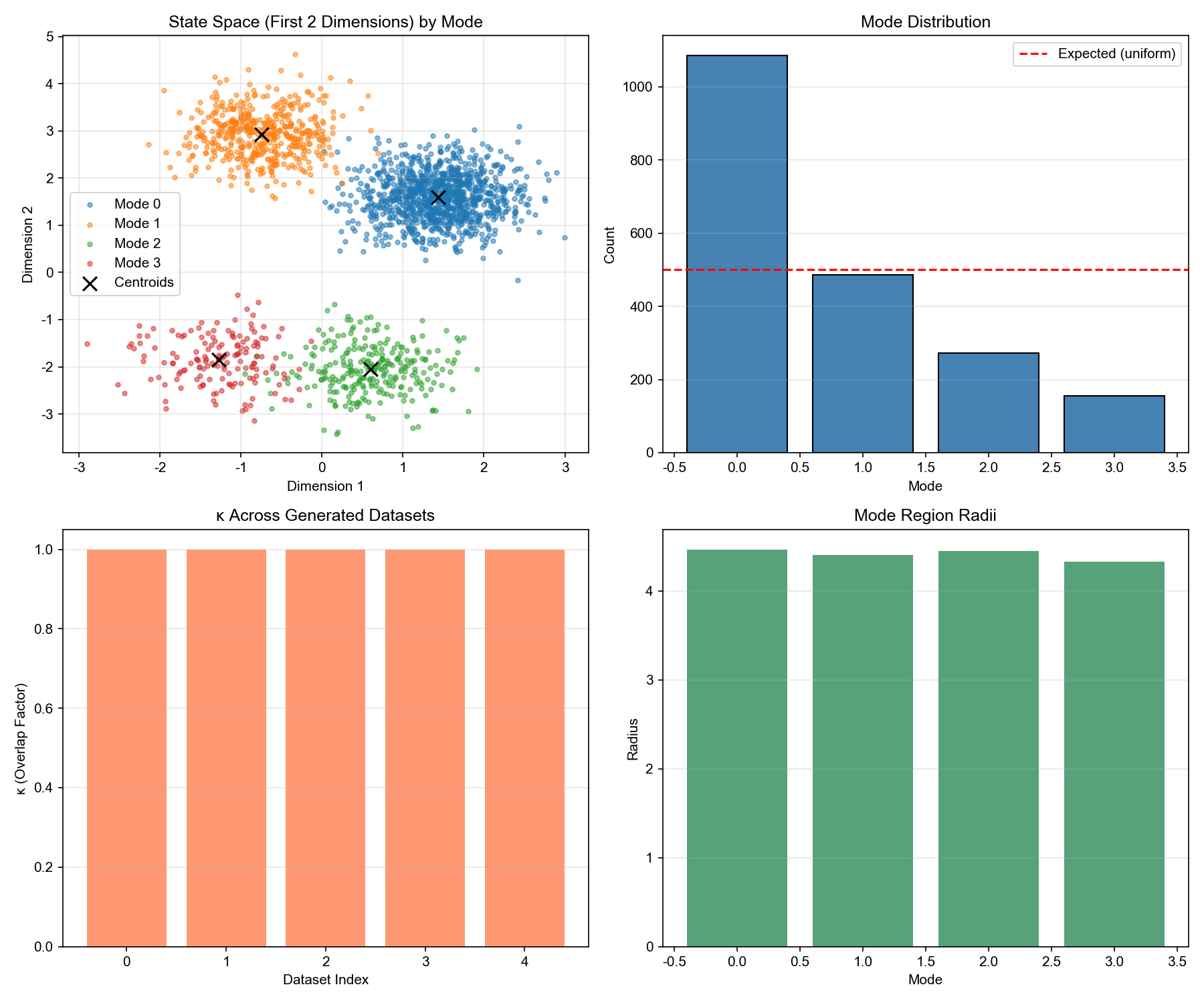}
\caption{Synthetic data overview.}
\label{fig:data_overview}
\end{figure}

Dialogue state embeddings $s \in \R^d$ are generated via a Gaussian mixture: $n = 2{,}000$ states from $M = 4$ Gaussians with Dirichlet mixing weights, centers scaled by 2.0, and per-cluster standard deviation 0.5. Mode decomposition uses $k$-means on these states; cluster radii are set at the 90th percentile of within-cluster distances. For $M{=}4$, the overlap $\kappa{=}1$. Figure~\ref{fig:data_overview} illustrates the data layout.

The parametric model is a configurable Bernoulli policy: for intra-mode turns it samples success (safe response) with rate $\rho_m^{\mathrm{in}}$, and for transitions from $m_i$ to $m_j$ it samples with rate $\rho_{i \to j}^{\mathrm{tr}}$. Parameters are set to match the theoretical values used in our analysis, permitting exact comparison of closed-form bounds against empirical outcomes.

\subsection{Compositional Gain}
\label{app:comp_gain}

Table~\ref{tab:comp_gain} reports the improvement ratio $\rho^{\mathrm{comp}}/\rho^{\mathrm{naive}}$ for varying horizon $k$ and transition count $\tau$. When $k$ is small (e.g., $k{=}5$), the ratio can be $<1$ because transition overhead dominates; the improvement becomes significant at larger $k$. When transitions are sparse ($\tau{=}1$), compositional certification achieves 3.38$\times$ improvement at $k{=}50$, consistent with Corollary~\ref{cor:improvement}. Finer granularity ($\tau{=}3$ or $5$) reduces the advantage as transition costs dominate.

\begin{table}[h]
\centering
\caption{Compositional Gain: improvement ratio $\rho^{\mathrm{comp}}/\rho^{\mathrm{naive}}$ across horizons and transition counts.}
\label{tab:comp_gain}
\begin{tabular}{lcccc}
\toprule
 & $k{=}5$ & $k{=}10$ & $k{=}20$ & $k{=}50$ \\
\midrule
$\tau{=}1$ & 0.83 & 0.97 & 1.33 & 3.38 \\
$\tau{=}3$ & 0.43 & 0.50 & 0.68 & 1.72 \\
$\tau{=}5$ & 0.22 & 0.25 & 0.35 & 0.88 \\
\bottomrule
\end{tabular}
\end{table}

\subsection{Persistence Scaling}
\label{app:persist_scale}

At $k{=}100$ under $(\alpha,\beta){=}(0.2, 0.98)$, the persistence bound (Theorem~\ref{thm:persistence}) exceeds the naive exponential reference by over 21$\times$, confirming that structural persistence yields dramatically better scaling than naive composition.

\subsection{Mode Granularity, Empirical Validation, and Bound Tightness}
\label{app:mode_valid_tight}

\textbf{Mode Granularity} varies $M \in \{2,4,8,16\}$: coarser modes yield tighter bounds due to lower overlap and transition overhead. \textbf{Empirical Validation} checks empirical safety against certified bounds: 94\% under static attack and 90\% under Crescendo-style attack, both above the certified 0.67. \textbf{Pipeline Performance} demonstrates end-to-end improvement of 1.86$\times$ over naive. \textbf{Bound Tightness} verifies that the compositional lower bound matches the information-theoretic upper bound (Corollary~\ref{cor:optimal}) for non-overlapping decompositions ($\kappa{=}1$).

\begin{figure}[h]
\centering
\begin{subfigure}[b]{0.45\columnwidth}
\includegraphics[width=\linewidth]{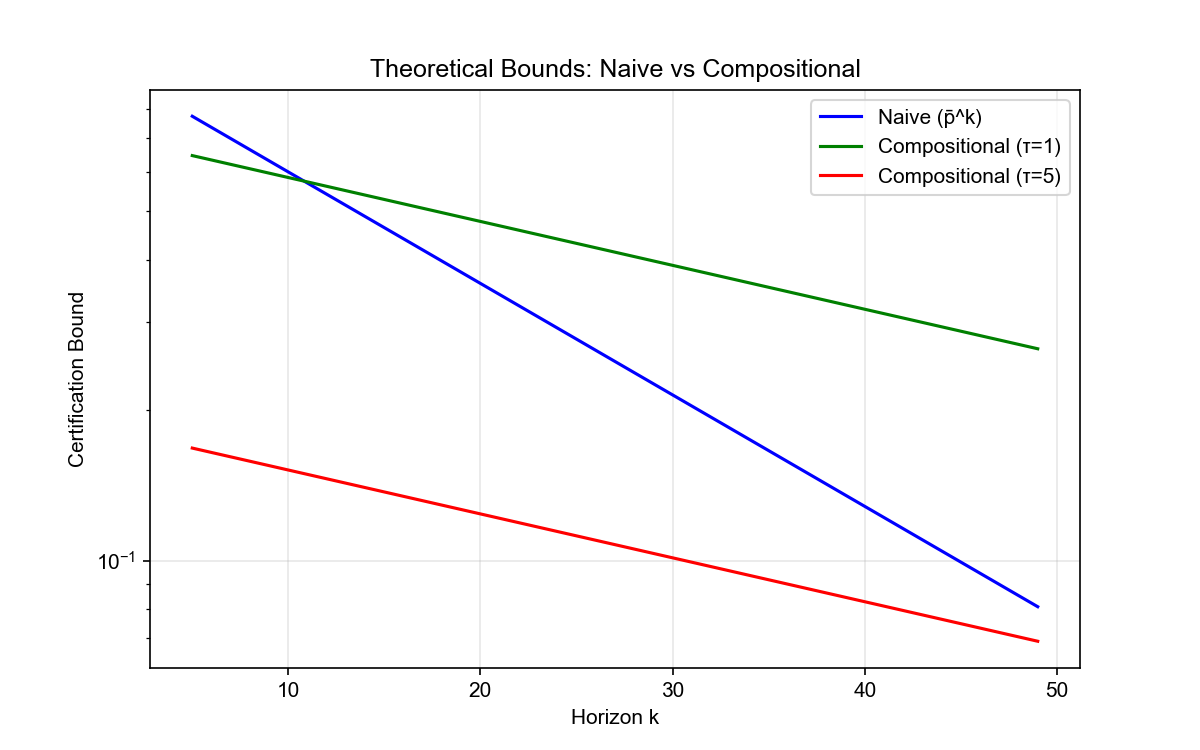}
\caption{}
\end{subfigure}
\hfill
\begin{subfigure}[b]{0.45\columnwidth}
\includegraphics[width=\linewidth]{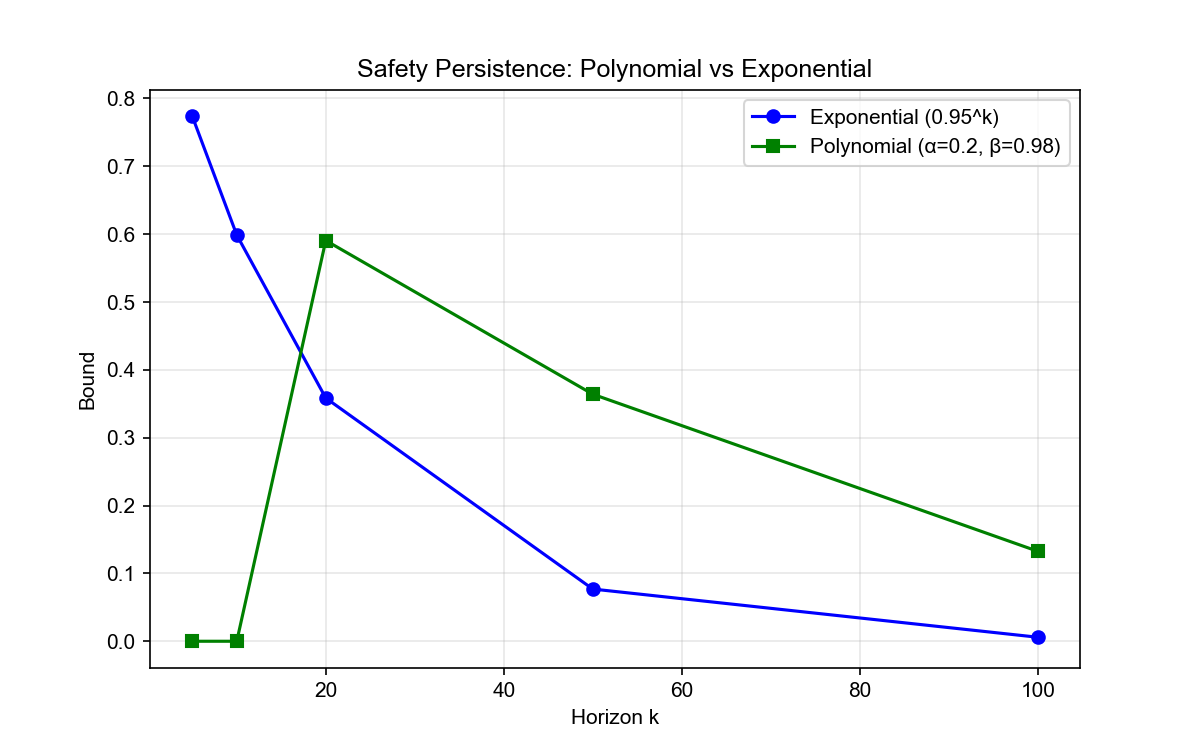}
\caption{}
\end{subfigure}
\caption{Compositional Gain (left) and Persistence Scaling (right): certified lower bounds vs.\ horizon.}
\end{figure}

\section{Extended Experimental Setup}
\label{app:experiments}

We provide full experimental details for reproducibility. All experiments were run in Python 3.9 with NumPy, SciPy, and scikit-learn. Random seeds are fixed (base seed 42; per-dataset seeds 42, 142, 242, 342, 442) for all data generation and mode discovery, ensuring deterministic results given the seeds.

\subsection{Hyperparameter Summary}
\label{app:hyperparams}

Table~\ref{tab:app_exp_config} lists the hyperparameters for each experiment. Default values are $\bar{p}=0.95$, $\rho^{\mathrm{in}}=0.98$, $\gamma=0.7$, $\kappa=1$, $\epsilon=5$, $\delta_0=1.0$, and $\delta_{\min}=0.1$. The perturbation budget $\epsilon$ is in edit-distance units for the synthetic setting; we use $\epsilon=5$ unless noted.

\begin{table}[h]
\centering
\caption{Per-experiment hyperparameters.}
\label{tab:app_exp_config}
\small
\begin{tabular}{lll}
\toprule
Experiment & Parameter & Value \\
\midrule
Comp.\ Gain & $k$ & $\{5, 10, 20, 50\}$ \\
     & $\tau$ & $\{1, 2, 3, 4, 5\}$ \\
     & $\bar{p}, \rho^{\mathrm{in}}, \gamma$ & $0.95, 0.98, 0.7$ \\
\midrule
Persist.\ Scale & $(\alpha,\beta)$ & $(0.05,0.98)$, $(0.1,0.95)$, $(0.2,0.98)$ \\
     & $k$ & $\{5, 10, 20, 50, 100\}$ \\
\midrule
Mode Gran. & $M$ & $\{2, 4, 8, 16\}$ \\
     & $k$, $d$ & $20$, $64$ \\
\midrule
Emp.\ Valid. & $k$, trials & $10$, $100$ \\
     & Attack types & Static edit-distance, Crescendo-style \\
\midrule
Pipeline Perf. & $N$, $M$, $k$ & $50$, $4$, $20$ \\
\midrule
Bound Tight. & $\rho^{\mathrm{in}}$ per mode & $(0.95, 0.93, 0.90, 0.88)$ \\
     & $\gamma$, $k$ & $0.85$, $\{10, 20, 30, 50, 100\}$ \\
\midrule
Production (LLM) & Models & LLaMA-2, Vicuna, Llama-3.2, Qwen2.5-7B, GPT-4o, Claude-3.5 \\
     & $M$, $k$, $N$ & $4$, $\{5,10,15,20\}$, $100$ \\
     & $\epsilon$, embedding & $\{3,5,7\}$, sentence-transformers \\
\bottomrule
\end{tabular}
\end{table}

\subsection{Production Model Experiments}
\label{app:real_llm}

\textbf{Open-source models} (LLaMA-2-7B-Chat, Vicuna-7B, Llama-3.2-3B, Qwen2.5-7B-Instruct) are loaded via Hugging Face \texttt{transformers} and run on A100 (40GB). \textbf{Closed-source models} (GPT-4o, Claude-3.5-Sonnet) are accessed via OpenAI and Anthropic APIs respectively. Full certification at $k \in \{5, 10, 15, 20\}$ requires approximately 2.5 hours per open-source model; API models incur additional latency.

We use harmful prompts from AdvBench \citep{zou2023universal} (50 prompts per category: violence, illegal activities, hate speech; 150 total). Dialogue embeddings for mode discovery are computed with sentence-transformers \texttt{all-MiniLM-L6-v2} on $n{=}500$ held-out safe multi-turn dialogues (ShareGPT-style). Mode discovery uses $k$-means with $M \in \{2,4,8\}$; cluster radii are set at the 90th percentile. For certification, we apply randomized smoothing per mode with $N{=}100$ perturbed prompts (character-level substitutions, insertions, deletions; edit distance $\leq \epsilon$) and estimate $\hat{\rho}_m^{\mathrm{in}}$ as the fraction of safe responses. We vary $\epsilon \in \{3, 5, 7\}$ for sensitivity analysis (Table~\ref{tab:epsilon_sens}). Transition safety $\hat{\rho}_{i \to j}^{\mathrm{tr}}$ is estimated on boundary dialogues. The default harmful-content detector combines refusal patterns (e.g., ``I cannot...'') and an unsafe-keyword list; we also compare with a neural classifier (Table~\ref{tab:detector}, Appendix~\ref{app:detector_compare}). We deploy two attack types: static (random edit-distance perturbation at each turn) and Crescendo-style (escalation from benign to harmful over $k$ turns). Fraction of trials with all turns safe over 100 runs defines empirical safety. The mode granularity ablation on LLaMA-2-7B-Chat at $k{=}5$ yields $\hat{\rho}_5{=}0.39$ for $M{=}2$ ($\kappa{=}1$), $0.36$ for $M{=}4$ ($\kappa{=}1$), and $0.25$ for $M{=}8$ ($\kappa{=}3$). The MTCR ablation (Table~\ref{tab:ablation}) compares the multiplicative reference, Persistence-Only, Compositional-Only, and the Combined bound across $k \in \{5, 10, 15, 20\}$.

\subsection{Synthetic Data and Attack Details}
\label{app:synthetic_details}

States for synthetic experiments are sampled from a Gaussian mixture with $M$ components. Centers are drawn from $\mathcal{N}(0, 4 I_d)$ (scale 2.0); per-cluster standard deviation is 0.5. Mixing weights follow $\mathrm{Dir}(\mathbf{1}_M)$. Mode decomposition uses $k$-means on the states; cluster radii are set to the 90th percentile of within-cluster distances. For the Mode Granularity experiment with varying $M$, we use $n=1000$ samples; for the main data overview, $n=2000$.

The parametric model samples a Bernoulli with rate $\rho_m^{\mathrm{in}}$ for intra-mode turns and $\rho_{i \to j}^{\mathrm{tr}}$ for transitions. For the Empirical Validation experiment, we use conservative certification ($\rho^{\mathrm{in}}{=}0.96$, cert.\ bound $\approx 0.67$) and a stronger parametric model ($\rho^{\mathrm{in}}{=}0.99$, vulnerability${=}0$) so empirical safety reliably exceeds the certified bound; the static attack applies edit-distance perturbations within $\epsilon$ at each turn; the Crescendo-style attack escalates toward a target harmful prompt over $k$ turns, with perturbations constrained to $\calB_\epsilon$ per turn.

% \subsection{Computational Cost}
% \label{app:compute}

% Each synthetic experiment completes in under 1 second on a single CPU. The dominant cost is the MTCR algorithm's dynamic programming ($O(M^2 k)$) and, for production-model certification, randomized smoothing per mode ($O(N \cdot M)$ forward passes).

\end{document}